\documentclass[11pt]{article}

\usepackage[]{coling}

\usepackage{times}
\usepackage{latexsym}

\usepackage[T1]{fontenc}
\usepackage[utf8]{inputenc}

\usepackage{microtype}

\usepackage{inconsolata}

\usepackage{graphicx}

\usepackage{times}
\usepackage{latexsym}
\usepackage{enumitem}
\usepackage[normalem]{ulem}
\usepackage{microtype}
\usepackage{hyperref}
\usepackage{url}
\usepackage{booktabs}
\usepackage{multirow}
\usepackage{inconsolata}
\usepackage{graphicx}
\usepackage{todonotes}
\usepackage{titlesec}
\usepackage{xcolor}
\usepackage{arydshln}

\title{Better to Ask in English: Evaluation of Large Language Models on English, Low-resource and Cross-Lingual Settings}

\author{ \bf
Krishno Dey$^a$, 
Prerona Tarannum$^b$,  
Md. Arid Hasan$^a$,  \\ 
\textbf{Imran Razzak}$^c$, 
\textbf{Usman Naseem}$^d$\\
{\small \{krishno.dey, arid.hasan\}@unb.ca}\\
$^a$University of New Brunswick, $^b$Daffodil International University, \\$^c$University of New South Wales, $^d$Macquarie University
}

\begin{document}
\maketitle
\begin{abstract}
Large Language Models (LLMs) are trained on massive amounts of data, enabling their application across diverse domains and tasks. Despite their remarkable performance, most LLMs are developed and evaluated primarily in English. Recently, a few multi-lingual LLMs have emerged, but their performance in low-resource languages, especially the most spoken languages in South Asia, is less explored. To address this gap, in this study, we evaluate LLMs such as GPT-4, Llama 2, and Gemini to analyze their effectiveness in English compared to other low-resource languages from South Asia (e.g., Bangla, Hindi, and Urdu). Specifically, we utilized zero-shot prompting and five different prompt settings to extensively investigate the effectiveness of the LLMs in cross-lingual translated prompts. The findings of the study suggest that GPT-4 outperformed Llama 2 and Gemini in all five prompt settings and across all languages. Moreover, all three LLMs performed better for English language prompts than other low-resource language prompts. This study extensively investigates LLMs in low-resource language contexts to highlight the improvements required in LLMs and language-specific resources to develop more generally purposed NLP applications.
\end{abstract}

\section{Introduction}
\label{introduction}
%recent status of LLMs
Large Language Models (LLMs) have recently undergone significant advancements and have transformed the landscape of Natural Language Processing (NLP). LLMs are trained using large datasets, which enable them to recognize, translate, predict, or generate text and other content \cite{yang2023harnessing}. Recent advances in LLMs enable the development of more powerful, efficient, and versatile NLP systems with broad applicability across different domains and tasks \cite{zhao2023survey}. 

%use of LLMs

Researchers have been evaluating LLMs for several NLP tasks, including machine translation \cite{xu2023paradigm, lyu2023new}, text summarization \cite{pu2023summarization, zhang2023summit}, reasoning \cite{suzgun2022challenging, miao2023selfcheck}, and mathematics \cite{lu2023mathvista, rane2023enhancing}. These works laid the foundation for exploring LLMs for downstream tasks in low-resource languages. LLMs are predominantly used in high-resource languages to evaluate different NLP tasks \cite{xing2024designing}. Recently, there has been a few benchmarks that have utilized LLMs for low-resource languages \cite{aggarwal2022indicxnli, asai2023buffet}. However, they do not extensively study the languages that we are evaluating in this study. 

%Most of these works that utilized LLMs in these domains are limited to high-resource languages such as English \cite{xing2024designing}.Nonetheless, these works laid the foundation for exploring LLMs for downstream tasks in low-resource languages.

On the other hand, the prevailing literature predominantly employs traditional machine learning or transformer-based models for analyzing low-resource languages \cite{hedderich2020survey}. 
%However, despite their remarkable performance, most models are developed and evaluated primarily in English. 
Recently, a few multi-lingual LLMs have emerged, but their performance in low-resource languages (especially the most spoken languages in South Asia) is not as good as in English \cite{asai2023buffet}. 
%especially the most spoken languages in South Asia, and is less explored \cite{asai2023buffet}. 
Three of the most widely spoken languages in South Asia are Hindi, with 610 million speakers; Bangla, with 273 million speakers; and Urdu, with 232 million speakers. Moreover, Hindi is the 3\textsuperscript{rd}, Bangla is the 7\textsuperscript{th}, and Urdu is the 10\textsuperscript{th} most spoken language in the world. Despite collectively representing over 1 billion speakers, computational resources for these three languages remain limited. In this study, we explore the effectiveness of LLMs in Bangla, Hindi, and Urdu languages vs English\footnote{We chose these languages because the authors are native speakers of Bangla, Hindi, and Urdu.}. 

%Very few works in low-resource languages utilize LLMs to solve Natural Language Processing (NLP) tasks, and most of the work is limited to some extent. 
A very few noteworthy works employ LLMs in Bangla \cite{liu2023knowdee, hasan2023zero, aggarwal2022indicxnli}, Urdu \cite{koto2024zero, aggarwal2022indicxnli}, and Hindi \cite{kumar2021sentiment, koto2024zero, aggarwal2022indicxnli}. These studies show that LLMs can achieve identical and better (in some cases) results compared to transformer-based models and traditional machine-learning techniques. However, more intensive studies are required to evaluate LLMs in these low-resource languages to determine their effectiveness and suggest potential improvement required in LLMs.

To address this gap, in this study, we analyze various LLMs (e.g., GPT-4, Llama 2, and Gemini) with zero-shot learning across both English and low-resource languages (e.g., Bangla, Hindi, and Urdu) under different prompt settings. Generic LLM names are used throughout the paper, exact versions are GPT-4,  Llama-2–70b-chat, and Gemini Pro. Specifically, we evaluate their effectiveness in English compared to other low-resource languages from South Asia. We perform cross-lingual prompt translation from originally designed prompts to develop various prompt settings and investigate the performance of LLMs under those prompt settings. Our results show that GPT-4 outperformed Llama 2 and Gemini in all languages. Furthermore, Llama 2 cannot recognize prompts in low-resource languages, resulting in the lowest performance. Additionally, LLMs perform better in English than in other languages such as Bangla, Hindi, and Urdu. The key contributions of this work include:

%findings
\begin{itemize}
    % \item We design five prompt settings to investigate the effectiveness of LLMs on cross-lingual translated prompts. The experimental results suggest that the translation of a specific language does not greatly affect the prediction capabilities of LLMs. Moreover, LLMs sometimes produce better performance for translated prompts than for original prompts.

    \item We investigate the effectiveness of zero-shot prompting with LLMs (GPT-4, Llama 2, Gemini) for cross-lingual tasks involving low-resource languages from South Asia (Bangla, Hindi, Urdu) on XNLI and SIB-200 datasets. The findings demonstrate that the translation quality of prompts has minimal impact on LLM performance, suggesting potential for language-agnostic prompting techniques in low-resource NLP applications.

    \item Our work explores five unique prompt settings to analyze the influence of prompt design on LLM performance in high-resource (English) and low-resource languages. The results show that English prompts consistently outperform prompts in other languages, even with translated prompts. 
    
    % This highlights the need for further research on prompt engineering specifically for low-resource language contexts.
    
    % We design five prompt settings to investigate the effectiveness of LLMs on cross-lingual translated prompts. The experimental result suggests the translation of a specific language does not affect the prediction capabilities of LLMs to a great extent. Moreover, LLMs sometimes produce better performance for translated prompts than original prompts.
    % \item We analyze and investigate the effectiveness of LLMs on English and other low-resource languages across five prompt settings. The outcome of the investigation suggests that it is better to prompt in English, as LLMs provide better responses for English prompts even when translated from other languages.
    
    \item We present a novel approach incorporating Natural Language Inference (NLI) and zero-shot prompting. Providing task descriptions and expected outputs through NLI creates a richer context for LLMs, improving performance in English and low-resource language tasks. 

    \item We conduct a comparative analysis of 3 prominent LLMs (GPT-4, Llama 2, Gemini) to evaluate their effectiveness in low-resource languages. The findings reveal that GPT-4 outperforms the other two models across all languages and prompt settings.
    
    % This information provides valuable insights for researchers and developers working on improving LLM performance in low-resource NLP settings.
    
    % This approach has the potential to improve the generalizability of LLMs for cross-lingual NLP applications.

    % \item 
    
    % We analyze and investigate the effectiveness of LLMs on English and other low-resource languages across five prompt settings. The outcome of the investigation suggests it is better to ask in English, LLMs provide better responses for English prompts even when translated from other languages.
    
    % \item We use natural language instructions (NLI) along with zero-shot prompting, incorporating task descriptions and expected outputs. This approach generates an appropriate context for LLMs to generate more suitable outputs.
    
    % We use natural language instructions along with zero-shot prompting, incorporating task descriptions and expected outputs. This approach generates an appropriate context for LLMs to generate more suitable outputs.
    
\end{itemize}

%The rest of the article is organized as follows: section \ref{sec:related-works} presents the summary of the current literature. Section \ref{sec:methodology} outlines our research methodology and experimental design. We discuss our experimental result in section \ref{sec:results-and-discussion}. In section \ref{sec:limitations} we highlight the limitations of our study. Finally, in Section \ref{sec:conclusion-future-work}, we conclude our study with a proposal for future research to improve upon our current work.
\section{Related Works}
\label{sec:related-works}
%Large Language Models (LLMs) are a significant advent that has a prominent milestone in the field of Natural Language Processing(NLP).

The emergence of LLMs marks a major milestone in NLP, proficient in various tasks such as language understanding~\cite{dou2019investigating}, text generation~\cite{huang2020towards}, question answering~\cite{giampiccolo2007third}, and sentiment analysis~\cite{yu2018diverse}. However, their performance in low-resource languages needs improvement \cite{robinson2023chatgpt}.

% The emergence of LLMs represents a major advancement and marks a significant milestone in the field of NLP. LLMs are proficient across several NLP downstream tasks such as language understanding~\cite{dou2019investigating}, text generation~\cite{huang2020towards}, question answering~\cite{giampiccolo2007third}, sentiment analysis~\cite{yu2018diverse}. However, LLM's performance across several tasks in low-resource languages is still limited and has significant room for improvement \cite{robinson2023chatgpt}. 
%Below we have discussed some previous works on LLMs for English and low-resource languages.

%Previous study \cite{kew2023turning} states that English-centric LLMs such as Llama 2 can almost perfectly match the input and output of the language when adjusted with few multilingual conversational instructions, despite their limited exposure to other languages. Another finding demonstrates, that with the widespread adoption of ChatGPT, LLMs can perform various tasks by following instructions and learning from just a few examples provided in context \cite{brown2020language}.

%the training data of GPT-3 is consisted with 93\% of English data, so there's only 7\% of other multilingual documents \cite{brown2020language}. 
%However, a strategy \cite{cui2023efficient} to improve the non-English languages with a downsampled dataset is used by training continuously.

\subsection{LLM for English}

In recent years, extensive resources and benchmarks for English \cite{wang2018glue, williams2017broad} have fueled the development of LLMs. LLMs significantly impact tasks such as question answering \cite{akter2023depth, tan2023can, zhuang2023toolqa}, reasoning \cite{suzgun2022challenging, miao2023selfcheck}, and machine translation \cite{xu2023paradigm, lyu2023new}. They are also applied in NLI \cite{gubelmann2023truth}, Sentiment Analysis \cite{sun2023sentiment, zhang2023sentiment}, and Hate Speech Detection \cite{zhang2024don}. Multi-lingual LLMs show strong performance across languages \cite{sitaram2023everything}, yet struggle with low-resource languages \cite{ahuja2023mega}. However, LLMs are known for their capabilities to understand the relationships among text sequences and produce results similar to state-of-the-art techniques \cite{pahwa2023bphigh, gubelmann2023truth}. The study of \citet{brown2020language} shows that LLMs can perform various tasks by following instructions and learning from just a few examples provided in context. Moreover, English-centric LLMs such as Llama 2 can almost perfectly match the input and output of the language when adjusted with few multi-lingual conversational instructions, despite their limited exposure to other languages \cite{kew2023turning}. The study of \citet{asai2023buffet} focuses on few-shot learning and instruction fine-tuning of smaller LLMs (such as mT5, mT0) and ChatGPT to improve the tasks (such as NLI, question answering, sentiment, commonsense reasoning, etc.) performance.

\subsection{LLM for Low-resource Languages}

Research and practical applications in the field of NLP are centred around high-resource languages that typically have large annotated corpora, well-established tools and libraries, and robust language models trained on large data. Despite being resourceful, the researchers prioritize analyzing high-resource languages and overlook other languages spoken by billions of people \cite{bender2019benderrule}. 

%previous works use machine learning
Most of the work across several downstream tasks in low-resource languages employs traditional machine learning or transformer-based language models \cite{jahan2023systematic, chhabra2023literature}. To improve the Natural Language Interface (NLI) task across low-resource languages, several researchers have attempted to develop benchmark datasets and frameworks \cite{bhattacharjee2021banglabert, rahman2017framework, chakravarthy2020detecting}.  Similar attempts were also made in tasks such as Sentiment Analysis \cite{islam2021sentnob, hasan2020sentiment, hasan2023blp, patra2018sentiment, muhammad2023innovations}, Hate Speech Detection\cite{badjatiya2017deep, zimmerman2018improving}, etc. These works do not explore the use of LLMs for those tasks above. 

%recent works what LLMs
Researchers have recently been shifting their attention towards developing applications for low-resource languages. As a result, several authors are evaluating LLMs in low-resource languages across several tasks \cite{liu2023knowdee, hasan2023zero, kabir2023benllmeval, koto2024zero, kumar2021sentiment, hee2024recent, garcia2023leveraging}. However, there are several challenges in evaluating LLMs in low-resource languages for different tasks \cite{ahuja2023mega, chung2023instructtods}. Researchers have developed low-resource language benchmark \cite{asai2023buffet} and datasets \cite{aggarwal2022indicxnli, khan2024indicllmsuite} to address such challenges. INDICXNLI \cite{aggarwal2022indicxnli}, INDICLLMSUITE \cite{khan2024indicllmsuite} dataset consists of several Indian regional languages but do not have any resources in Urdu. The benchmark BUFFET \cite{asai2023buffet} incorporates most spoken South Asian languages in NLI tasks; however, they do not incorporate Bangla in their experiments. Along with the several low-resource benchmarks and datasets, many prompting techniques were also explored to evaluate LLMs on low-resource languages \cite{qin2023chain-of-thought, huang2023not-all-language}. 

%What are we doing? 
Recently, a few multi-lingual LLMs have emerged, but their performance in low-resource languages, especially the most spoken languages in South Asia, is less explored. To address this gap, in this study, we evaluate LLMs such as GPT-4, Llama 2, and Gemini to analyze their effectiveness in English compared to other low-resource languages from South Asia (e.g., Bangla, Hindi, and Urdu).

\section{Methodology}
\label{sec:methodology}

\begin{table*}[!t]\vspace{-0.5cm}
\centering
\small
% \resizebox{\linewidth}{!}{
\scalebox{0.80}{
\begin{tabular}{p{1.3cm}|p{12cm}}
\hline
\textbf{Model} & \textbf{Prompt Template} \\ \hline

\textbf{GPT-4} & 
[
\{ 

`role': `user',

`content': "Classify the following `premise' and `hypothesis' into one of the following classes: `Entailment', `Contradiction', or `Neutral'. Provide only label as your response."

premise: [PREMISE\_TEXT]

hypothesis: [HYPOTHESIS\_TEXT]

label:

\},

\{

role: `system',

content: "You are an expert data annotator and your task is to analyze the text and find the appropriate output that is defined in the user content."

\}
]
\\ \hline
\textbf{Llama 2 and Gemini} & 
Classify the following `premise' and `hypothesis' into one of the following classes: `Entailment', `Contradiction', or `Neutral'. Provide only label as your response.

premise: [PREMISE\_TEXT]

hypothesis: [HYPOTHESIS\_TEXT]

label:
\\
\bottomrule
\end{tabular}
}
\caption{Prompts used for zero-shot learning on XNLI dataset}
\label{tab:prompt-design-xnli}
\end{table*} 

%prompt approach
\subsection{Prompt Approach}
%why good prompt design is important?
The quality of the prompt impacts the performance of LLMs \cite{white2023prompt}. A well-designed prompt helps users and developers leverage the capabilities of LLMs to retrieve information that aligns with their specific interests. LLMs are usually trained on diverse datasets. Thus, providing clear instructions on interacting with LLMs and producing the desired information is essential. Designing a good prompt is an iterative process that requires instructions refining through successive interactions, enabling the LLMs to learn and produce desired results.
%prompt design in our study
In this study, we have employed zero-shot prompting and provided instruction in natural language. To help LLMs generate more relevant output, we provided detailed instructions containing the task descriptions and prompt template in Table \ref{tab:prompt-design-xnli} (and \ref{tab:prompt-design-sib} in the Appendix). We used the same instruction format for all languages and LLMs. %Table \ref{tab:prompt-design-xnli} presents an English prompt utilizing zero-shot learning with the Gemini model for a data sample.
%In this case, the model made a wrong prediction, the actual label was neutral, and however model predicted contradiction.  

%role in gpt4
Furthermore, to leverage the ability of GPT-4 to receive role information and perform according to that specific role, we provided role information along with the prompt. 
% safety setting in Gemini
During the experiment, Gemini Pro blocked most prompts for containing harmful and inappropriate texts and made no predictions. To obtain the predictions for those harmful contents, we changed the safety settings of the Gemini Pro models. Despite changing the safety setting, models still did not make any predictions. See {Appendix \ref{sec:appendix_a}  for safety settings.
% are available in.

%\begin{table}[!ht]
%\centering
%\small
%\begin{tabular}{p{1.1cm} p{5cm}p{3cm}p{3cm}}
%\toprule
%\textbf{M} & \textbf{P} & \textbf{AL} & \textbf{PL} \\ 
%\midrule   
%\textbf{Gemini} & 
%Classify the following `premise' and `hypothesis' into one of the following classes: `Entailment', `Contradiction', or `Neutral'. Provide only label as your response. 

%premise: [PREMISE\_TEXT]

%hypothesis: [HYPOTHESIS\_TEXT]

%label: & Neutral & Contradiction\\
%\midrule
%\multicolumn{4}{p{12cm}}{Here, \textbf{PREMISE\_TEXT:} The doors were locked when we went in. \textbf{HYPOTHESIS\_TEXT:} We had the keys with us.}\\
%\bottomrule
%\end{tabular}
%\caption{Sample prompt and response for English prompt employing Gemini. M: Model, P: Prompt, AL: Actual Label, PL: Predicted Label}
%\label{tab:sample-prompt-and-response}
%\end{table}

\begin{table*}[!t]
\centering
% \resizebox{1\linewidth}{!}{
\scalebox{0.63}{
\begin{tabular}{llrr|rr|rr|rr|rr}
\toprule
\multirow{2}{*}{\textbf{Model}}& \multirow{2}{*}{\textbf{Lang.}} & \multicolumn{2}{c}{\textbf{P1}}& \multicolumn{2}{c}{\textbf{P2}}& \multicolumn{2}{c}{\textbf{P3}}& \multicolumn{2}{c}{\textbf{P4}}& \multicolumn{2}{c}{\textbf{P5}}\\ 
\cmidrule{3-12} 
&& \textbf{Acc.} & \textbf{F1$_{macro}$} & \textbf{Acc.} & \textbf{F1$_{macro}$} & \textbf{Acc.} & \textbf{F1$_{macro}$} & \textbf{Acc.} & \textbf{F1$_{macro}$}  & \textbf{Acc.} & \textbf{F1$_{macro}$} \\ 
\midrule

\multirow{4}{*}{\textbf{GPT-4}} 
        & \textbf{BN} & 68.72 & \textcolor{blue}{69.05}  & 68.72	& \textcolor{blue}{69.05}   & - & -   & 70.73 & \textcolor{blue}{71.18}   & 70.26 & \textcolor{blue}{70.54}\\
        & \textbf{EN} & 86.73 & \textcolor{blue}{\underline{86.79}} & - & -   & 87.03 & \textcolor{blue}{\underline{87.08}}   & 82.42 & \textcolor{blue}{\underline{81.99}}   & 86.73 & \textcolor{blue}{\underline{86.81}}\\
        & \textbf{HI} & 71.52 &	\textcolor{blue}{71.97} & 70.26	& \textcolor{blue}{70.73}  & 68.52 & \textcolor{blue}{68.80}   & 71.20 & \textcolor{blue}{71.67}   & - & -\\ 
        & \textbf{UR} & 65.07 &	\textcolor{blue}{64.77} & 66.45	& \textcolor{blue}{66.73}   & 65.31 & \textcolor{blue}{65.48}   & - & -   & 66.77 & \textcolor{blue}{66.95}\\ \midrule

\multirow{4}{*}{\textbf{Llama 2}} 
        & \textbf{BN} & 35.49 & 36.73 & 33.24	& 31.27   & - & -   & 35.42 & 35.91   & 32.05 & 33.59\\
        & \textbf{EN} & 65.73 & \underline{58.76} & - & -   & 60.80	& \underline{56.78} & 62.59 & \underline{59.17}   & 63.59 & \underline{60.11}\\
        & \textbf{HI} & 36.27 & 38.57  & 33.53 & 33.66   & 38.02	& 40.89   & 33.19 & 33.43   & - & -\\ 
        & \textbf{UR} & 36.39 & 35.56  & 39.48 & 35.83   & 36.77 & 37.65   & - & -   & 33.31 & 	36.35 \\ \midrule

\multirow{4}{*}{\textbf{Gemini}} 
        & \textbf{BN} & 62.19 & 61.74 & 60.56	& 60.16   & - & -   & 56.56 & 52.41   & 60.67 & 60.30\\
        & \textbf{EN} & 73.71 & \underline{73.36} & - & -   & 73.65	& \underline{73.09}   & 62.94 & \underline{60.73}   & 74.71 & \underline{74.72}\\
        & \textbf{HI} & 61.90 & 61.72   & 61.88 & 61.70   & 62.10	& 61.47   & 52.10 & 48.19   & - & -\\ 
        & \textbf{UR} & 49.75 & 46.11 & 57.30 & 56.47   & 59.04 & 58.40   & - & -   & 57.25 & 	56.34 \\

\bottomrule
\end{tabular}
}
\caption{Performances of the LLMs across the settings, and languages for XNLI dataset. \underline{Underline} shows the best F1 score among four languages across all settings and LLMs. \textcolor{blue}{Blue} indicates the best F1 score in all settings among three LLMs. %\textbf{Bold} represents the highest F1 score among LLMs in all five prompt settings for each of the four languages. 
Lang.: Language, Acc.: Accuracy, BN: Bangla, EN: English, HI: Hindi, and UR: Urdu}
\label{tab: xnli-performance}
\end{table*}
%data

%data and tasks

\subsection{Experimental Details}

\subsubsection{Data}

In this study, we employed LLMs to evaluate the NLI and Classification task due to the absence of cross-lingual datasets for other downstream NLP tasks.\\
% Why are we conducting the study on NLI task and the only available dataset
%We are conducting our study solely on the NLI task due to the absence of a cross-lingual dataset for other downstream NLP tasks.
We used the most widely known and publicly available NLI datasets, the cross-lingual natural language inference (XNLI) \cite{conneau2018xnli} and SIB-200 \cite{adelani2024sib200}. %XNLI corpus extends the Multi-Genre NLI (MultiNLI) with raw text from Open American National Corpus. 
\noindent \textbf{XNLI:} The dataset contains 15 languages, including low-resource languages such as Urdu and Hindi. English validation and test sets of MultiNLI were manually translated, while the train set was machine-translated for all languages. The dataset contains $392,702$ train, $2,490$ validation, and $5,010$ test samples. Each data sample contains the premise, hypothesis, and corresponding labels (e.g., entailment, neutral, and contradiction). In our study, we selected test sets of English, Hindi, and Urdu languages from the XNLI dataset. We also selected a Bangla XNLI dataset available for public use, and %was developed by \cite{bhattacharjee2021banglabert}. This dataset 
was generated by translating the XNLI dataset using English to Bangla translator model \cite{hasan2020not}, which is widely used and the only large-scale study on English-to-Bangla machine translation. The Bangla XNLI dataset contains $381,449$ train, $2,419$ validation, and $4,895$ test samples.% (i.e. the Bangla XNLI dataset has 115 fewer test samples than the XNLI test set). 
%The table \ref{tab:data-distribution} provides class-wise test set distribution. 

\noindent \textbf{SIB-200:} We used SIB-200 dataset for the classification task. SIB-200 is an inclusive and big evaluation dataset that contains more than 200 languages \cite{adelani2024sib200}. The dataset is straightforward and mostly used for classification tasks. The dataset has 701 train samples, 99 validation samples, and 204 test samples. SIB-200 has been developed from the machine translation corpus Flores-200 \cite{nllbteam2022language}. Later, the dataset was extended to 203 languages with sentence-level annotation.

%Why we selected only Bangla, English, Hindi, and Urdu (South Asian language, Huge Population, and Native speakers)
The most spoken languages in South Asia—Hindi, Bangla, and Urdu are used daily by approximately 1.1 billion people \footnote{\url{https://en.wikipedia.org/wiki/List\_of\_languages\_by\_total\_number\_of\_speakers}}. Despite their widespread use, these languages are considered low-resource in NLP. We included all three in the XNLI dataset to assess LLMs' performance compared to English. Having at least one native speaker of these languages among the authors aids in organizing and understanding LLMs' responses during experiments.

% The most widely spoken languages in South Asia are Hindi, Bangla, and Urdu, with approximately $1.1$ billion people using these languages daily \footnote{\url{https://en.wikipedia.org/wiki/List\_of\_languages\_by\_total\_number\_of\_speakers}}. Despite being spoken by nearly $1/8th$ world population, these three languages are still among the most low-resource languages in the NLP domain. We selected all three South Asian languages (i.e., Hindi, Bangla, and Urdu) from the XNLI dataset to evaluate the LLMs generalization across these low-resource languages compared to English. Additionally, the author list contains at least one native speaker of these languages, which is advantageous during the experiment for organizing and understanding the prompts and responses of LLMs.

% \begin{table}[!ht]
% \centering
% \begin{tabular}{lllr}
% \toprule
% \multicolumn{1}{c}{\textbf{Task}} & \multicolumn{1}{c}{\textbf{Languages}} & \multicolumn{1}{c}{\textbf{Class}} & \multicolumn{1}{c}{\textbf{Test}}\\ \midrule

% \multirow{6}{*}{\textbf{NLI}} & \multirow{3}{*}{\textbf{EN, HI, UR}} & Contradiction & $1,670$ \\
% & & Entailment & $1,670$ \\
% & & Neutral & $1,670$ \\ \cline{2-4}
% & \multirow{3}{*}{\textbf{BN}} & Contradiction & $1,630$ \\
% & & Entailment & $1,631$ \\
% & & Neutral & $1,634$ \\
% \bottomrule
% \end{tabular}
% \caption{Class-wise test set data distribution for all the tasks. EN: English, BN: Bangla, HI: Hindi, and UR: Urdu.} 
% \label{tab:data-distribution}
% \end{table}

%Experimental Settings
\subsubsection{Prompt Settings}
Table \ref{tab:prompt-setting-xnli} in the Appendix shows the prompt settings and prompts of all languages used in our study for the XNLI dataset. Setting P1 consists of four language-specific prompts, and each of the four prompts was designed by 4 native speakers of Bangla, English, Hindi, and Urdu languages. Setting P2, P3, P4, and P5 was created by translating the prompt from English, Bangla, Urdu, and Hindi respectively. Classes names such as 'Entailment', 'Contradiction', and 'Neutral' were used consistently for all language prompts \footnote{Class labels are in English for all languages}. Table  \ref{tab:prompt-setting-sib} in the Appendix shows a similar prompt setting used for the SIB-200 dataset.
%In setting P2, the original English prompt from settings P1 was translated into Bangla, Hindi, and Urdu prompts. Similarly, in the P3 Setting, the original Bangla prompt from setting P1 was translated into English, Hindi, and Urdu. Furthermore, in Setting P4, we translated the Urdu prompt from the P1 setting prompt to Banlga, English, and Hindi prompts. In setting P5, the original Hindi prompt from P1 was translated into Banlga, English, and Urdu prompts. 
We used Google Translator%\footnote{\url{https://translate.google.com/}} 
for translating prompts across four languages. 

%P3, P4, and P5 we translated the original prompt of each of the four languages (BN, EN, HI, UR) to the other three languages (i.e. In setting P2, we translated the EN prompt to BN, HI, and UR prompts and similarly, in setting P3 we translated BN prompt to EN, HI, and UR prompt). 

%post processing 
\subsubsection{Post-Processing}
\label{post-processing}
Our designed prompt instructed the LLMs to output the class labels (e.g., entailment, neutral, and contradiction) as the response. However, LLMs often returned extra characters and words along with the label. We processed the outputs by removing unknown characters and filtering the class labels using regular expressions to handle such responses. When the LLM response contained no class label, we classified them as ``None" to get the final results. %(e.g. if the actual label of an instance is ``contradiction" we classified it as ``entailment") to generate the final results.

%evaluation metrics
\subsubsection{Evaluation Metrics}

To evaluate the performances, we computed accuracy, precision, recall, and F1 scores for all the experimental settings to evaluate the effectiveness of LLMs on cross-lingual translated prompts. We have computed weighted precision, recall, and macro F1 scores to deal with class imbalance.

\section{Results and Discussion}
\label{sec:results-and-discussion}
In this section, we present and discuss the experimental results of our study. Firstly, we discuss the performance of LLMs in English compared to low-resource languages (i.e., Bangla, Hindi, and Urdu). Then, we provide a comparison of LLMs across different settings. Finally, we delve into the results of each setting (e.g., P1, P2, P3, P4, and P5). Additionally, we provide class-wise precision and recall scores in Table \ref{tab:class-wise-performance-xnli} and \ref{tab:class-wise-performance-sib}. %Table \ref{tab: performance} illustrates the performance across models and languages for all five settings. Table \ref{tab:class-wise-performance} detailed class-wise F1$_{macro}$ score. Class-wise precision and recall scores are presented in Appendix \ref{sec:appendix_b}.

\begin{table*}[!t]
\centering
% \resizebox{1\linewidth}{!}{
\scalebox{0.63}{
\begin{tabular}{llrr|rr|rr|rr|rr}
\toprule
\multirow{2}{*}{\textbf{Model}}& \multirow{2}{*}{\textbf{Lang.}} & \multicolumn{2}{c}{\textbf{P1}}& \multicolumn{2}{c}{\textbf{P2}}& \multicolumn{2}{c}{\textbf{P3}}& \multicolumn{2}{c}{\textbf{P4}}& \multicolumn{2}{c}{\textbf{P5}}\\ 
\cmidrule{3-12} 
&& \textbf{Acc.} & \textbf{F1$_{macro}$} & \textbf{Acc.} & \textbf{F1$_{macro}$} & \textbf{Acc.} & \textbf{F1$_{macro}$} & \textbf{Acc.} & \textbf{F1$_{macro}$}  & \textbf{Acc.} & \textbf{F1$_{macro}$} \\ 
\midrule

\multirow{4}{*}{\textbf{GPT-4}} 
        & \textbf{BN} & 86.27 & \textcolor{blue}{84.35}   & 86.76	& \textcolor{blue}{85.00}   & - & -   & 85.78 & \textcolor{blue}{83.80}   & 86.76 & \textcolor{blue}{84.77}\\
        
        & \textbf{EN} & 86.27 & \textcolor{blue}{84.35} & - & -   & 87.75 & \textcolor{blue}{\underline{85.45}}  & 88.72 & \textcolor{blue}{\underline{86.63}}   & 87.74 & \textcolor{blue}{\underline{85.52}}\\
        
        & \textbf{HI} & 84.80 &	\textcolor{blue}{83.01} & 87.74	& \textcolor{blue}{
        86.72}  & 86.76 & \textcolor{blue}{85.25}   & 85.78 & \textcolor{blue}{83.29}   & - & -\\ 
        
        & \textbf{UR} & 86.76 &	\textcolor{blue}{\underline{84.46}} & 86.76	& \textcolor{blue}{84.03}  & 85.78 & 83.02   & - & -   & 74.51 & \textcolor{blue}{68.61}\\ \midrule

\multirow{4}{*}{\textbf{Llama 2}} 
        & \textbf{BN} & 25.33 & 17.22 & 40.00	& 11.43   & - & -   & 24.29	& 5.91   & 12.86 & 4.87\\
        & \textbf{EN} & 58.97 & \underline{57.42} & - & -   & 71.43	& \underline{67.48} & 50.00 & \underline{46.54}   & 69.88 & \underline{66.68}\\
        & \textbf{HI} & 25 & 8.20 & 34.97 & 22.79   & 27.09	& 10.25   & 23.53 & 7.86   & - & -\\ 
        & \textbf{UR} & 19.89 & 5.14  & 25.00 & 5.76   & 23.76 & 5.59   & - & -   & 23.64 & 	5.67 \\ \midrule

\multirow{4}{*}{\textbf{Gemini}} 
        & \textbf{BN} & 84.80 & \underline{83.93} & 85.29	& 83.79   & - & -   & 82.35 & 80.80   & 84.80 & 82.57\\
        & \textbf{EN} & 84.80 & 82.66 & - & -   & 83.82	& 81.75   & 82.84 & 80.19   & 85.29 & \underline{83.10}\\
        & \textbf{HI} & 83.33 & 79.75   & 87.25 & 85.85   & 82.35	& 79.09   & 83.82 & \underline{81.80}   & - & -\\ 
        & \textbf{UR} & 81.37 & 77.86 & 81.37	& 79.17  & 86.76 & \textcolor{blue}{\underline{84.10}}   & - & -   & 73.04 & 67.05 \\

\bottomrule
\end{tabular}
}
\caption{Performances of the LLMs across the settings, and languages for SIB-200 dataset. \underline{Underline} shows the best F1 score among four languages across all settings and LLMs. \textcolor{blue}{Blue} indicates the best F1 score in all settings among three LLMs. %\textbf{Bold} represents the highest F1 score among LLMs in all five prompt settings for each of the four languages.
Lang.: Language, Acc.: Accuracy, BN: Bangla, EN: English, HI: Hindi, and UR: Urdu}
\label{tab:sib-performance}
\end{table*}

%English vs low-resource languages
\subsection{English vs Low-resource Languages}

The experimental results presented in Table \ref{tab: xnli-performance} and \ref{tab:sib-performance} indicate that all LLMs exhibit superior performance for English prompts (indicated by \underline{Underline}) in most of the prompt settings. Note that the P2 setting does not contain English prompts. %To ensure fair comparisons across different languages and prompt settings, we consider the macro F1 score. 
In the P1 setting (original prompts), GPT-4 outperforms other languages in English by $17.75$\%, $14.82$\%, and $22.02$\% for Bangla, Hindi, and Urdu respectively for the XNLI dataset. This trend persists across all settings, with English consistently demonstrating superior performance compared to other languages, even with translated prompts. Similarly, for the SIB-200 dataset, GPT-4 prompts in all languages produce better performance than other languages. However, the performance of GPT on the SIB-200 dataset is more balanced compared to XNLI for prompts in all languages. The smaller size of the SIB-200 dataset could contribute to such a balanced performance.    
%In settings P3 and P4, GPT-4 exhibits enhanced performance in English compared to Hindi and Urdu by $18.28$\% and $21.60$\%, respectively. Similarly, in setting P5, English outperforms Bangla and Urdu by $10.81$\% and $10.33$\%, respectively. 
Although Bangla and Hindi perform better than Urdu, there remains a significant performance gap compared to English on the XNLI dataset. However, the performance differences among low-resource languages across all settings are very small.

Similarly, Llama 2 performs better in English but poorly in other languages. The performance gap between English and other languages is substantial in setting P1, with English outperforming Bangla, Hindi, and Urdu by $22.03$\%, $20.19$\%, and $23.20$\%, respectively for the XNLI dataset. This trend persists across all settings, with English consistently performing better than other languages. In the SIB-200 dataset, the performance gap between English and the other three languages in the P1 setting is $40.2$, $49.22$, and $52.28$ for Bangla, Hindi, and Urdu respectively. The English language prompts produce better performance than other languages for all other settings.    

In contrast, the performance disparity between English and other languages is slightly lower in Gemini Pro compared to GPT-4 and Llama 2. Gemini Pro performs better in English than Bangla, Hindi, and Urdu by $11.62$\%, $11.64$\%, and $27.25$\%, respectively in P1 settings for the XNLI dataset. This trend continues in translated prompt settings (P3, P4, and P5), with English consistently outperforming other languages. For the SIB-200 dataset, English prompts produce poor performance across all settings except for setting P5. However, the difference margin in performance among languages is very small and similar. 

Such bias towards English in LLMs can be attributed to several factors. One significant reason is the disproportionate training distribution, with most LLMs predominantly trained on extensive English corpora. For instance, in Llama 2, approximately $90$\% of the training data is sourced from English, leaving only $10$\% representing other languages worldwide. This skewed training distribution significantly contributes to the observed biases towards low-resource languages. The findings underscore the need for substantial improvements in LLMs to ensure better generalization across low-resource languages. The balanced performance of LLMs on the SIB-200 dataset could be because of several factors. The SIB-200 dataset is significantly smaller compared to the XNLI dataset. The smaller size may have contributed to the increased quality of the SIB-200 dataset, making it easier for LLMs to understand and respond.

\begin{table*}[!t]
\centering
%\def\arraystretch{0.5}
%\resizebox{\linewidth}{6cm}{%
%\resizebox{\linewidth}{!}{
\scalebox{0.55}{
\begin{tabular}{lllrr|rrr|rrr|rrr|rrr}
\toprule
\multirow{2}{*}{\textbf{Model}} & \multirow{2}{*}{\textbf{Lang}}& \multicolumn{3}{c}{\textbf{P1}}& \multicolumn{3}{c}{\textbf{P2}}& \multicolumn{3}{c}{\textbf{P3}}& \multicolumn{3}{c}{\textbf{P4}}& \multicolumn{3}{c}{\textbf{P5}}\\ 
\cmidrule{3-17} 
& &\textbf{Cont.}& \textbf{Ent.} & \textbf{Neut.}  &  \textbf{Cont.}& \textbf{Ent.} & \textbf{Neut.} &  \textbf{Cont.}& \textbf{Ent.} & \textbf{Neut.} &  \textbf{Cont.}& \textbf{Ent.} & \textbf{Neut.} &  \textbf{Cont.}& \textbf{Ent.} & \textbf{Neut.} \\ 
\midrule

\multirow{4}{*}{\textbf{GPT-4}} 
        & \textbf{BN} &  73.51 & 66.72 & 66.91 & 73.51 & 66.72 & 66.91 & - & - & - & 76.81 & 69.93 & 66.80 & 74.92 & 68.59 & 68.11\\
                                        
        & \textbf{EN} & 90.90 & 87.56 & 81.92 & - & - & - & 91.22 & 87.84 & 82.17 & 89.83 & 83.44 & 72.71 & 90.91 & 87.72 & 81.80\\
                                        
        & \textbf{HI}& 77.85 & 69.81 & 68.24 & 76.97 & 67.96 & 67.26 & 76.06 & 63.88 & 66.46 & 78.30 & 69.62 & 67.08  & - & - & -\\
                                        
        & \textbf{UR} & 73.20 & 61.74 & 59.37& 72.68 & 63.47 & 64.03 & 71.57 & 61.04 & 63.82 & - & - & - & 73.47 & 63.32 & 64.06\\ \midrule

\multirow{4}{*}{\textbf{Llama 2}} 
        & \textbf{BN} & 06.39 & 11.91 & 91.88 &   09.30 & 39.02 & 45.48 &    -  &   -    &  -      &   10.90 & 35.74 & 61.08 &  03.70 & 28.87 & 68.19\\
                                        
        & \textbf{EN} & 88.36 & 68.11 & 19.79 &   -     &   -   &   -   &    76.77 & 61.38 & 32.20 &   82.97 & 62.59 & 31.95 &  80.62 & 63.16 & 36.56 \\
                                        
        & \textbf{HI}&  11.84 & 14.38 & 89.48 &   0.60  & 0.77  & 99.61 &    18.29 & 24.40 & 79.99 &   00.90 & 00.66 & 98.02 &  - & - & -\\
                                        
        & \textbf{UR} & 29.61 & 40.41 & 36.65 &   39.02 & 45.60 & 22.86 &    24.32 & 37.55 & 51.09 &   -     &   -   &  -    &  11.00 & 12.64 & 85.42\\ \midrule

\multirow{4}{*}{\textbf{Gemini}} 
        & \textbf{BN} & 70.08 & 59.88 & 55.25 &  69.03 & 54.74 & 56.72 &  - & - & -             &  68.98 & 30.09 & 58.17 &  68.63 & 53.30 & 58.98 \\
                                        
        & \textbf{EN} & 82.05 & 72.53 & 65.48 &  - & - & -             &  80.61 & 74.51 & 64.14 &  79.23 & 42.43 & 60.53 &  82.53 & 73.92 & 67.72\\
                                        
        & \textbf{HI} & 70.37 & 57.10 & 57.68 &  69.67 & 56.08 & 59.34 &  70.57 & 61.40 & 52.43 &  63.41 & 25.01 & 56.15 &  - & - & -             \\
                                        
        & \textbf{UR} & 53.75 & 20.95 & 63.62 &  67.05 & 50.56 & 51.79 &  68.84 & 50.56 & 55.80 &  - & - & -             &   66.62 & 51.16 & 51.24\\

\bottomrule
\end{tabular}
}
\caption{ Class-wise F1$_{macro}$ score for GPT-4, Llama 2, and Gemini across five prompt settings for the XNLI dataset. Lang.: Language, BN: Bangla, EN: English, HI: Hindi, and UR: Urdu, Cont: contradiction, Ent: Entailment, Neut: Neutral.}
\label{tab:class-wise-performance-xnli}
\end{table*}

\begin{table*}[!t]
\centering
\resizebox{\linewidth}{!}{
\begin{tabular}{lllrrr|rrrr|rrrr|rrrr|rrrr}
\toprule
\multirow{2}{*}{\textbf{Model}} & \multirow{2}{*}{\textbf{Class}}& \multicolumn{4}{c}{\textbf{P1}}& \multicolumn{4}{c}{\textbf{P2}}& \multicolumn{4}{c}{\textbf{P3}}& \multicolumn{4}{c}{\textbf{P4}}& \multicolumn{4}{c}{\textbf{P5}}\\ 
\cmidrule{3-22} 
& &\textbf{BN}& \textbf{EN} & \textbf{HI} & \textbf{UR} &\textbf{BN}& \textbf{EN} & \textbf{HI} & \textbf{UR} &
\textbf{BN}& \textbf{EN} & \textbf{HI} & \textbf{UR} &
\textbf{BN}& \textbf{EN} & \textbf{HI} & \textbf{UR} &
\textbf{BN}& \textbf{EN} & \textbf{HI} & \textbf{UR} \\ 
\midrule

\multirow{4}{*}{\textbf{GPT-4}} 
        & \textbf{Ent.} 
        &  85.71 & 80.00 &	75.00 & 70.97 %--- P1
        & 82.35 & - & 81.25 & 75.00 %--- P2
        & - & 82.35 & 75.00 & 66.67 %--- p3
        & 78.79 & 82.35 & 78.79 & -             %--- p4
        & 80.00 & 82.35 & - & 68.75 \\  %--- p4
        
        & \textbf{Geo.} 
        &  76.47 & 74.29 &	71.79 & 81.08 
        & 77.78 & - & 78.95 & 68.75 
        & - & 74.29 & 77.78 & 81.08
        & 80.00 & 74.29 & 80.00 & -
        & 77.78 & 74.29 & - & 52.83\\

        & \textbf{Hel.} 
        & 85.0 & 75.68 & 80.95 & 77.78 
        & 75.68 & - & 83.72 & 85.00 
        & - & 76.92 & 82.05 & 76.92
        & 73.68 & 82.05 & 73.68 & -
        & 76.92 & 75.68 & - & 75.68\\
                                        
        & \textbf{Pol.}
        & 91.53 & 95.08 & 93.33 & 91.53 
        & 91.53 & - & 94.92 & 94.92
        & - & 95.08 & 93.33 & 91.53
        & 89.66 & 95.08 & 89.66 & -
        & 91.53 & 95.08 & - & 23.53\\
        
        & \textbf{Sci.} 
        & 94.34 & 88.89 & 88.68 & 88.50 
        & 90.74 & - & 88.46 & 93.46
        & - & 91.74 & 88.68 & 90.09
        & 90.91 & 92.59 & 90.91 & -
        & 91.59 & 90.09 & - & 86.96 \\

        & \textbf{Spr.} 
        & 89.80 & 89.80 & 86.96 & 89.80 
        & 90.20 & - & 91.67 & 85.71
        & - & 87.50 & 93.88 & 8571
        & 87.50 & 89.80 & 87.5 & -
        & 87.50 & 89.80 & - & 86.79\\

        & \textbf{Tra.} 
        & 87.06 & 86.75 & 84.34 & 91.57 
        & 86.75 & - & 88.10 & 85.39 
        & - & 90.24 & 86.05 & 89.16
        & 86.05 & 90.24 & 86.05 & -
        & 88.10 & 91.36 & - & 85.71\\

        \midrule

\multirow{4}{*}{\textbf{Llama 2}}

        & \textbf{Ent.} 
        &  0.00 & 64.52 & 08.70 & 0.00 
        & 0.00 & - & 09.52 & 0.00 
        & - & 77.78 & 10.53 & 0.00
        & 0.00 & 40.00 & 0.00 & -
        & 0.00 & 72.00 & - & 0.00\\
        
        & \textbf{Geo.} 
        &  17.39 & 42.55 & 0.00 &  0.00
        & 0.00 & - & 09.52 & 0.00 
        & - & 46.81 & 0.00 & 0.00
        & 0.00 & 34.29 & 0.00 & -
        & 0.00 & 46.51 & - & 0.00\\

        & \textbf{Hel.} 
        & 09.09 & 56.25 & 0.00 &  0.00
        & 0.00 & - & 25.00 & 0.00 
        & - & 66.67 & 0.00 & 0.00
        & 0.00 & 29.63 & 08.70 & -
        & 0.00 & 75.00 & - & 0.00\\
                                        
        & \textbf{Pol.}
        & 27.03 & 68.97 & 0.00 & 0.00
        & 0.0 & - & 16.67 & 0.00 
        & - & 80.60 & 06.25 & 0.00
        & 0.00 & 57.14 & 0.0 & -
        & 0.00 & 80.70 & - & 0.00\\
        
        & \textbf{Sci.} 
        & 39.69 & 64.71 & 39.83 & 36.00 
        & 57.14 & - & 45.28 & 40.32
        & - & 83.05 & 41.32 & 13.64
        & 41.35 & 53.68 & 38.14 & -
        & 34.11 & 78.10 & - & 39.67\\

        & \textbf{Spr.} 
        & 22.22 & 66.67 & 0.00 & 0.00 
        & 0.00 & - & 07.14 & 0.00 
        & - & 87.50 & 0.00 & 0.0
        & 0.00 & 64.86 & 0.0 & -
        & 0.00 & 89.47 & - & 0.00\\

        & \textbf{Tra.} 
        & 05.13 & 38.30 & 08.89 & 0.00
        & 0.00 & - & 46.43 & 0.00 
        & - & 30.00 & 13.64 & 0.0
        & 0.00 & 46.15 & 08.16 & -
        & 0.00 & 25.00 & - & 0.00\\

        \midrule
\multirow{4}{*}{\textbf{Gemini}} 

        & \textbf{Ent.} 
        &  72.73 & 81.25 & 76.47 & 59.26 
        & 80.00 & - & 78.79 & 72.73 
        & - & 82.35 & 56.25 & 75.00
        & 73.68 & 77.42 & 72.73 & -
        & 81.08 & 81.25 & - & 60.61\\
        
        & \textbf{Geo.} 
        &  76.47 & 68.97 & 53.33 & 58.82 
        & 77.78 & - & 78.79 & 64.71
        & - & 73.33 & 64.52 & 64.71
        & 70.59 & 68.97 & 68.97 & -
        & 66.67 & 64.52 & - & 55.00\\

        & \textbf{Hel.} 
        & 84.21 & 75.68 & 78.95 & 81.08
        & 76.92 & - & 87.18 & 75.68
        & - & 70.27 & 84.21 & 87.18
        & 75.00 & 66.67 & 81.08 & -
        & 76.92 & 82.05 & - & 79.07\\
                                        
        & \textbf{Pol.}
        & 96.67 & 88.89 & 90.32 & 88.52 
        & 91.53 & - & 88.52 & 87.10 
        & - & 87.50 & 88.89 & 91.80
        & 93.33 & 90.32 & 88.89 & -
        & 92.06 & 86.57 & - & 27.78\\
        
        & \textbf{Sci.} 
        & 83.64 & 86.44 & 90.91 & 85.22
        & 88.29 & - & 90.27 & 84.48
        & - & 86.18 & 86.24 & 90.91
        & 87.76 & 84.30 & 85.71 & -
        & 89.72 & 88.50 & - & 82.88\\

        & \textbf{Spr.} 
        & 92.31 & 90.57 & 86.79 & 90.20 
        & 86.27 & - & 90.57 & 88.46 
        & - & 87.27 & 90.20 & 90.20
        & 87.50 & 86.79 & 92.31 & -
        & 88.00 & 92.31 & - & 77.97\\

        & \textbf{Tra.} 
        & 81.48 & 86.84 & 81.48 &  81.93
        & 85.71 & - & 86.84 & 81.08
        & - & 85.33 & 83.33 & 88.89
        & 77.78 & 86.84 & 92.93 & -
        & 83.54 & 86.49 & - & 86.05\\

\bottomrule
\end{tabular}
}
\caption{ Class-wise F1$_{macro}$ score for GPT-4, Llama 2, and Gemini across five prompt settings for the SIB-200 dataset. Lang.: Language, BN: Bangla, EN: English, HI: Hindi, and UR: Urdu, Ent: Entertainment, Geo: Geography, Hel: Health, Sci: Science/Technology, Spr: Sports, Tra: Travel.}
\label{tab:class-wise-performance-sib}
\end{table*}

Detailed class-wise F1 scores are provided in Table \ref{tab:class-wise-performance-xnli} and Table~\ref{tab:class-wise-performance-sib}. The class-wise result is balanced for both GPT-4 and Gemini Pro for the XNLI dataset. In contrast, Llama 2 is biased towards the Neutral class. This behavior may be due to its inability to fully comprehend the prompt, leading to random predictions in the Neutral class without proper analysis. Similarly, for the SIB-200 dataset, GPT-4 and Gemini Pro produce balanced class-wise results and are biased toward the "Science/Technolog" class for Llama 2. The intensity of bias is higher in the SIB-200 dataset, where Llama 2 could not generate any prediction for many classes across all settings. 

In summary, GPT-4 demonstrated superior performance to Llama 2 and Gemini Pro across all settings and languages, showcasing its accuracy in understanding prompts and providing appropriate responses. Gemini Pro performed better than Llama 2 but lacked support for Urdu in the XNLI dataset and struggled with predicting samples containing harmful content. Conversely, Llama 2 faced challenges in understanding prompts in low-resource languages. These findings suggest that GPT-4's multi-lingual capabilities and precise, prompt understanding contribute to its effectiveness in various language settings.

\subsection{Cross-Lingual Translated Prompts}

This section discusses the results of LLMs across languages in each setting. To make valid comparisons among different settings (i.e., P1, P2, P3, P4, and P5), we only consider the best F1 score achieved among the three LLMs (i.e., GPT-4, Llama 2, and Gemini Pro) for each language. Table \ref{tab: xnli-performance} and \ref{tab:sib-performance} show the best scores for each language for individual prompt settings (indicated by \textcolor{blue}{Blue}).

%For the original prompts in the P1 setting, the highest F1 scores for each language are Bangla (69.05\%), English (86.79\%), Hindi (71.97\%), and Urdu (64.77\%). In P2, the highest F1 scores for each language are Bangla (69.05\%), Hindi (70.73\%), and Urdu (66.73\%). For setting P3, the highest F1 scores for each language are English (87.08\%), Hindi (68.80\%), and Urdu (65.48\%). In setting P4, the highest F1 scores for each language are Bangla (71.18\%), English (81.99\%), and Hindi (71.67\%). Finally, for setting P5, the highest F1 scores for each language are Bangla (86.81\%), English (70.54\%), and Urdu (66.95\%).

In the XNLI dataset, the Hindi-to-Bangla translated prompt in setting P5 achieved a higher F1 score (86.81\%) than the original Bangla prompt in setting P1 (69.05\%). Similarly, for the Urdu language, the translated prompt in setting P5 achieved a higher F1 score (66.95\%) than the original Urdu prompt. For the English language, the Bangla-to-English translated prompt in setting P3 achieved a higher F1 score (87.08\%) than the original English prompt. However, the original prompt achieved the highest F1 score for the Hindi language compared to other translated prompts. Similarly, in the SIB-200 dataset setting P5 has a better F1 score than the original Bangla prompt. Translated prompts in settings P4 and P3 have better F1 scores than original English and Hindi prompts. However, the original Urdu prompt has a better F1 score compared to any translated prompts.    
Our results suggest that %translating a prompt to another language could improve the quality of the prompt. Moreover, 
the translation of a specific language does not significantly affect the prediction capabilities of LLMs. %Furthermore, the performance of LLMs is sometimes better for translated prompts than original prompts in setting P1.

\subsection{Comparison among LLMs}

We compare the performance of different LLMs on XNLI and SIB-200 datasets. Results showed that GPT4 and Llama 2 could make predictions for all data samples, while Gemini Pro did not make any predictions for prompts containing harmful content despite adjusting the safety setting. The number of unpredicted samples was very low (ranging from 1-4) and negligible. Additionally, LLMs returned unknown characters and words alongside class labels, which we addressed during post-processing (described in \ref{post-processing}). We assigned an inverse class to samples with invalid labels, reducing the overall performance of LLMs during evaluation metric calculations. See Table~\ref{tab: miss-classification-xnli} and \ref{tab: miss-classification-sib} in Appendix for the total number of invalid labels returned by LLMs.

% is shown in Table~\ref{tab: miss-classification-xnli} and Table~\ref{tab: miss-classification-sib} in Appendix.

% \begin{table}[!t]
% \centering
% % \resizebox{\linewidth}{!}{
% \scalebox{0.60}{
% \begin{tabular}{llrrrrr}
% \toprule
% \multirow{1}{*}{\textbf{Model}} & \multirow{1}{*}{\textbf{Lang.}} &\multicolumn{1}{c}{\textbf{ P1}} & \textbf{ P2}& \textbf{ P3}& \textbf{ P4} & \textbf{ P5}\\ 
% %\cmidrule{3-7} 
% %&& \multicolumn{1}{c}{\textbf{Miss}} & \multicolumn{1}{c}{\textbf{Miss.}} & \multicolumn{1}{c}{\textbf{Miss.}} & \multicolumn{1}{c}{\textbf{Miss.}} & \multicolumn{1}{c}{\textbf{Miss.}} \\ 
% \midrule

% \multirow{4}{*}{\textbf{GPT-4}} 
%         & \textbf{BN} & 1 &  13 & - & 2 & 48 \\
%         &\textbf{EN} & 0 &  - &  0 & 0 & 0 \\
%         &\textbf{HI} & 9 &  13 &  17  & 7 & -\\ 
%         &\textbf{UR} & 43 &  53 &  18  & - & 130\\ \midrule

% \multirow{4}{*}{\textbf{Llama 2}} 
%         &\textbf{BN} & 4147 & 1678 & - & 2195 & 2827\\
%         &\textbf{EN} & 19 & - & 619 & 322 & 588\\
%         &\textbf{HI} & 4010 & 4966 & 3306 & 4923  & -\\ 
%         &\textbf{UR} & 1308 & 790 & 1917 & - & 3872 \\ \midrule

% \multirow{4}{*}{\textbf{Gemini}} 
%         &\textbf{BN} & 142 & 149 & - & 68 & 127\\
%         &\textbf{EN} & 138 & - & 143 & 80 & 129\\
%         &\textbf{HI} & 111 & 131  & 138 & 133 & -\\ 
%         &\textbf{UR} & 1330 & 105 & 85 & - & 77\\ 

% \bottomrule
% \end{tabular}
% }
% \caption{Number of invalid labels returned by LLMs across the settings, models, and languages for the XNLI dataset. Lang.: language.}
% \label{tab: miss-classification-xnli}
% \end{table}

Table \ref{tab: xnli-performance} and Table~\ref{tab:sib-performance} demonstrate that GPT-4 consistently outperforms Llama 2 and Gemini Pro across all five settings (indicated by \textcolor{blue}{Blue}) for both the datasets except for setting P3 on SIB-200 dataset. Further investigation shows that GPT -4 accurately understood prompts and provided appropriate responses, with minimal unwanted characters and words accompanying the labels. Interestingly, GPT-4 showed superior performance in English prompts across all settings, suggesting its robustness in understanding English language prompts compared to others. Moreover, GPT-4's multi-lingual capabilities improved performance in languages such as Bangla, Hindi, and Urdu.

Conversely, Gemini Pro performed better than Llama 2 but not as well as GPT-4. Moreover, Gemini Pro produced the best performance for Urdu in setting P3 for the SIB-200 dataset. Despite its performance, Gemini Pro returned many unwanted characters, words, or invalid labels, which affected its overall performance. Additionally, Gemini Pro struggled with predicting samples containing harmful content and lacked support for the Urdu language, further limiting its performance.

In contrast, Llama 2 exhibited poor performance compared to GPT-4 and Gemini Pro. Trained on 90\% English data, Llama 2 faced challenges in understanding prompts in low-resource languages such as Bangla, Hindi, and Urdu. Additionally, Llama 2 returned more unwanted characters, words, or invalid labels, significantly impacting its performance during evaluation metric calculations. Moreover, Llama 2 showed dominance in predicting the \textit{neutral} class over \textit{entailment} and \textit{contradiction} classes, indicating its difficulty in accurately predicting sentences belonging to these classes in the XNLI dataset.
\section{Conclusion and Future Work}
\label{sec:conclusion-future-work}

%summary

In this study, we evaluated LLMs including GPT-4, Llama 2, and Gemini Pro across English and low-resource languages such as Bangla, Hindi, and Urdu, focusing on the NLI task due to dataset limitations. Despite recent advancements in LLMs, their performance on low-resource languages, particularly those spoken in South Asian countries, remains underexplored. Utilizing zero-shot prompting techniques in five different settings, our findings reveal GPT-4 consistently outperforms Llama 2 and Gemini Pro, with Gemini Pro showing better performance than Llama 2, particularly struggling with low-resource language prompts. Interestingly, English prompts yield superior responses compared to low-resource language prompts, and translating prompts from one language to another occasionally enhances performance. Future research should prioritize collecting low-resource datasets and developing resources for South Asian languages to improve LLM generalization, while incorporating techniques like few-shot prompting and developing cross-lingual datasets for various tasks would facilitate comprehensive evaluations on low-resource languages.

\section*{Limitations}
\label{sec:limitations}
%Only one dataset and one task
% In this study, we evaluated LLMs on a single dataset (i.e. XNLI) and on a single task (i.e. NLI), which may not accurately reflect the capabilities of LLMs across a broader range of tasks and datasets. The unavailability of cross-lingual datasets such as XNLI restricted us from conducting our study on a broader scale incorporating more tasks and datasets. 
% % Only Zeroshot prompting was used. 
%  Additionally, we only utilized zero-shot prompting techniques and did not explore explicit prompting techniques (e.g. few-shot prompting) to enhance the performance of LLMs. Utilization of only prompting techniques may limit the generalization of our findings. However, we were unable to explore other prompting techniques due to resource limitations, conducting experiment with LLMs requires premium access and premium comes at a hefty cost. 

We evaluated LLMs on a single task (i.e., NLI), which may not accurately reflect the capabilities of LLMs across a broader range of tasks and datasets. The unavailability of cross-lingual datasets such as XNLI and SIB-200, restricted us from conducting our study on a broader scale, incorporating more tasks and datasets. Additionally, we only utilized zero-shot prompting techniques and did not explore explicit prompting techniques (e.g., few-shot prompting) to enhance the performance. The utilization of only prompting techniques may limit the generalization of our findings. However, we could not explore other prompting techniques due to resource limitations; conducting experiments with LLMs requires premium access, which comes at a hefty cost.

% In this study, we evaluated LLMs on a single dataset (i.e., XNLI) and a single task (i.e., NLI), which may not accurately reflect the capabilities of LLMs across a broader range of tasks and datasets. The unavailability of cross-lingual datasets such as XNLI restricted us from conducting our study on a broader scale, incorporating more tasks and datasets. Additionally, we only utilized zero-shot prompting techniques and did not explore explicit prompting techniques (e.g., few-shot prompting) to enhance the performance of LLMs. The utilization of only prompting techniques may limit the generalization of our findings. However, we were unable to explore other prompting techniques due to resource limitations; conducting experiments with LLMs requires premium access, and premium access comes at a hefty cost.

\bibliography{custom}

\appendix

\appendix
\section{Prompt and Safety Settings}
\label{sec:appendix_a}

In this section, we present the prompts utilized in the study. Table \ref{tab:prompt-design-sib} shows zero-shot learning prompts design used for the SIB-200 dataset. Consistency was maintained across all settings, with the only variation being the language of the prompt. Figure \ref{tab:prompt-setting-xnli} and \ref{tab:prompt-setting-sib} show the prompts used in our study for XNLI and SIB-200 datasets respectively. 
%In this section, we present the safety settings in 
Table \ref{tab:safety_setting} that are used for Gemini Pro across all prompt settings to prevent it from blocking predictions for harmful content. 
%Table \ref{tab:safety_setting} shows the safety settings used for Gemini Pro across all prompt settings.  

%\begin{figure}[!t]
%\centering
%\begin{tabular}{ccc}
%\includegraphics[width=0.45\textwidth]{figures/setting-p1.png} &
%\includegraphics[width=0.48\textwidth]{figures/setting-p2.png} & \\
%\textbf{(P1)}  & \textbf{(P2)}\\[6pt]
%\end{tabular}
%\begin{tabular}{ccc}
%\includegraphics[width=0.48\textwidth]{figures/setting-p3.png} &
%\includegraphics[width=0.48\textwidth]{figures/setting-p4.png} \\
%\includegraphics[width=0.48\textwidth, height=3.5cm]{figures/setting-p5.png} \\
%\textbf{(P3)}  & \textbf{(P4)}  \\[6pt]
%\end{tabular}
%\begin{tabular}{cc}
%\includegraphics[width=0.48\textwidth]{figures/setting-p5.png} \\
%\textbf{(P5)}  \\[6pt]
%\end{tabular} %\textbf{P1, P2, P3, P4, and P5 } represent five different prompt setting used in this study
%\caption{Settings \textbf{(P1)} shows the four original prompts For Bangla, English, Hindi, and Urdu.   
%\textbf{(P2)} Setting shows the translated prompts from the original English prompt of setting P1.
%\textbf{(P3)} Setting shows the translated prompts from the Bangla prompt of setting P1.
%\textbf{(P4)} Setting shows the translated prompts from the Urdu prompt of setting P1.
%\textbf{(P5)} Setting shows the translated prompts from the Hindi prompt of setting P1. 
%BN: Bangla, EN: English, HI: Hindi, and UR: Urdu.
%}
%\label{fig:prompt-setting}
%\end{figure}

\begin{table}[!htpb]
\centering
% \resizebox{.99\linewidth}{!}{
\scalebox{0.70}{
\begin{tabular}{p{8cm}|p{2.5cm}}
\hline
\textbf{Category} & \textbf{Threshold} \\ \hline 
HARM\_CATEGORY\_HARASSMENT & BLOCK\_NONE \\
HARM\_CATEGORY\_HATE\_SPEECH & BLOCK\_NONE \\
HARM\_CATEGORY\_SEXUALLY\_EXPLICIT & BLOCK\_NONE \\
HARM\_CATEGORY\_DANGEROUS\_CONTENT & BLOCK\_NONE \\
HARM\_CATEGORY\_SEXUAL & BLOCK\_NONE \\
HARM\_CATEGORY\_DANGEROUS & BLOCK\_NONE \\
\bottomrule
\end{tabular}
}
\caption{Safety setting used for Gemini Pro model to prevent blocking the predictions for harmful content.}
\label{tab:safety_setting}
\end{table}

\begin{table*}[!]\vspace{-0.5cm}
\centering
\small
% \resizebox{\linewidth}{!}{
\scalebox{0.70}{
\begin{tabular}{p{1.3cm}|p{12cm}}
\hline
\textbf{Model} & \textbf{Prompt Template} \\ \hline

\textbf{GPT-4} & 
[
\{ 

`role': `user',

`content': "Classify the following 'text' into one of the following classes: 'science/technology', 'travel', 'politics', 'sports', 'health', 'entertainment', or 'geography'. Provide only class as your response.."

text: [TEXT]

label:

\},

\{

role: `system',

content: "You are an expert data annotator and your task is to analyze the text and find the appropriate output that is defined in the user content."

\}
]
\\ \hline
\textbf{Llama 2 and Gemini} & 
Classify the following 'text' into one of the following classes: 'science/technology', 'travel', 'politics', 'sports', 'health', 'entertainment', or 'geography'. Provide only class as your response.

text: [TEXT]

label:
\\
\bottomrule
\end{tabular}
}
\caption{Prompts used for zero-shot learning on SIB-200 dataset}
\label{tab:prompt-design-sib}
\end{table*}

\setlength\dashlinedash{5pt}
\setlength\dashlinegap{5pt}
\setlength\arrayrulewidth{1pt}

\begin{table*}[!ht]
\centering
\scalebox{0.70}{
% \resizebox{1\linewidth}{!}{
\begin{tabular}{lp{15cm}}
\toprule
\multicolumn{2}{c}{\textbf{P1 Settings}} \\
\midrule
\includegraphics[width=1\textwidth]{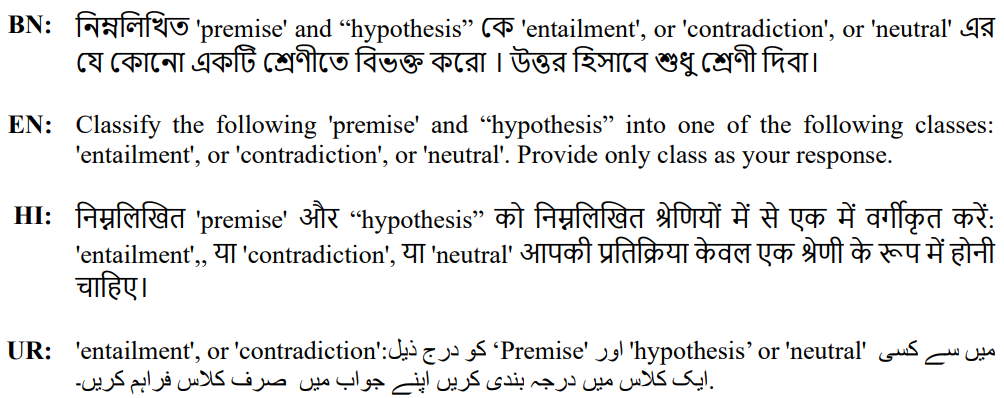} \\ \midrule

\multicolumn{2}{c}{\textbf{P2 Settings}} \\ \midrule
\includegraphics[width=1\textwidth]{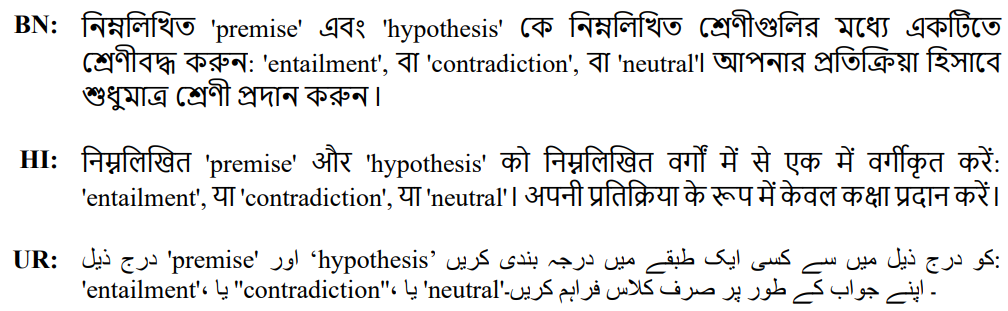} \\ \midrule

\multicolumn{2}{c}{\textbf{P3 Settings}} \\ \midrule
\includegraphics[width=1\textwidth]{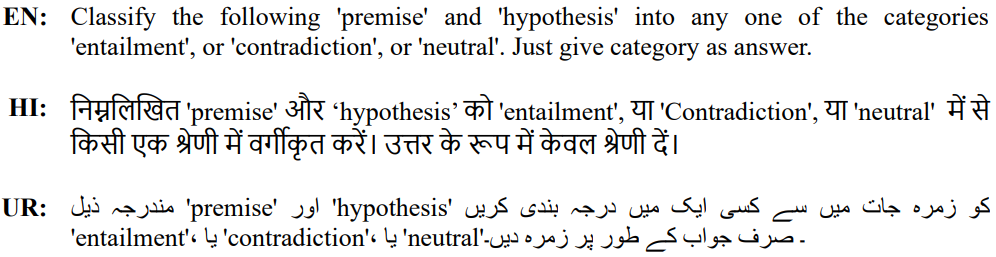} \\ \midrule

\multicolumn{2}{c}{\textbf{P4 Settings}} \\ \midrule
\includegraphics[width=1\textwidth]{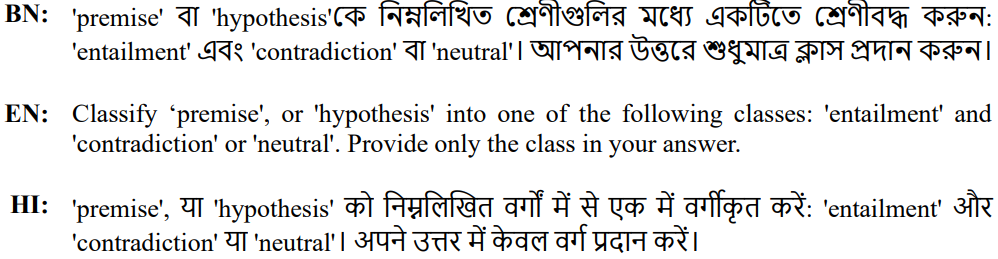} \\ \midrule

\multicolumn{2}{c}{\textbf{P5 Settings}} \\ \midrule
\includegraphics[width=1\textwidth]{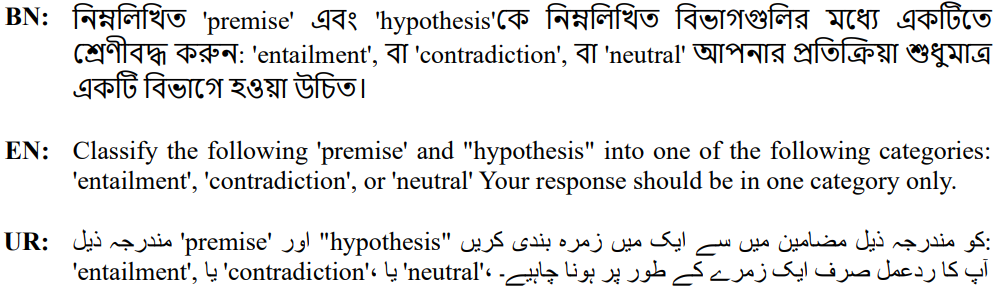}  \\ \midrule

\bottomrule
\end{tabular}
}
\caption{Settings \textbf{(P1)} represents prompts provided by native speakers for XNLI Dataset.   
\textbf{P2 Settings, P3 Settings, P4 Settings, and P5 Settings} represents the translated prompts from English, Bangla, Urdu, and Hindi respectively. English class names are used as labels in all datasets, as the class labels are in English for all languages.
BN: Bangla, EN: English, HI: Hindi, and UR: Urdu.
} 
\label{tab:prompt-setting-xnli}
\end{table*}

\setlength\dashlinedash{5pt}
\setlength\dashlinegap{5pt}
\setlength\arrayrulewidth{1pt}

\begin{table*}[!ht]
\centering
% \resizebox{1\linewidth}{!}{
\scalebox{0.70}{
\begin{tabular}{lp{15cm}}
\toprule
\multicolumn{2}{c}{\textbf{P1 Settings}} \\ \midrule
\includegraphics[width=1\textwidth]{figs/XNLI/setting-p1.png}\\ \midrule

\multicolumn{2}{c}{\textbf{P2 Settings}} \\ \midrule
\includegraphics[width=1\textwidth]{figs/XNLI/setting-p2.png} \\ \midrule

\multicolumn{2}{c}{\textbf{P3 Settings}} \\ \midrule
\includegraphics[width=1\textwidth]{figs/XNLI/setting-p3.png}\\ \midrule

\multicolumn{2}{c}{\textbf{P4 Settings}} \\ \midrule
\includegraphics[width=1\textwidth]{figs/XNLI/setting-p4.png} \\\midrule

\multicolumn{2}{c}{\textbf{P5 Settings}} \\ \midrule
\includegraphics[width=1\textwidth]{figs/XNLI/setting-p5.png} \\ \bottomrule
\end{tabular}
}
\caption{Settings \textbf{(P1)} represents prompts provided by native speakers for SIB-200 Dataset.   
\textbf{P2 Settings, P3 Settings, P4 Settings, and P5 Settings} represents the translated prompts from English, Bangla, Urdu, and Hindi respectively. English class names are used as labels in all datasets, as the class labels are in English for all languages.
BN: Bangla, EN: English, HI: Hindi, and UR: Urdu.
} 
\label{tab:prompt-setting-sib}
\end{table*}

\section{Detailed Experimental Results}
\label{sec:appendix_b}

This section presents the detailed experimental results for all three LLMs across five different prompt settings.
Table \ref{tab:detailed-result-gpt-llama-gemini-xnli} and \ref{tab:detailed-result-gpt-llama-gemini-sib} shows class-wise experimental results (precision and recall) for GPT-4, Llama 2, and Gemini Pro across different prompt settings. Table \ref{tab: miss-classification-xnli} and \ref{tab: miss-classification-xnli} show the total number of invalid labels returned by LLMs.

\begin{table*}[!]
\centering
% \resizebox{0.8\linewidth}{!}{
\scalebox{0.70}{
\begin{tabular}{llrrrrrrrrrr}
\toprule
 \multirow{2}{*}{\textbf{Lang.}}& \multirow{2}{*}{\textbf{Class}} & \multicolumn{2}{c}{\textbf{P1}}& \multicolumn{2}{c}{\textbf{P2}}& \multicolumn{2}{c}{\textbf{P3}}& \multicolumn{2}{c}{\textbf{P4}} & \multicolumn{2}{c}{\textbf{P5}}\\ \cmidrule{3-12}
 
 & & \textbf{P.} & \textbf{R.}  & \textbf{P.} & \textbf{R.} & \textbf{P.} & \textbf{R.} & \textbf{P.} & \textbf{R.} & \textbf{P.} & \textbf{R.}  \\
 
\midrule
\multicolumn{12}{c}{\textbf{GPT-4}} \\ 
\midrule

\multirow{3}{*}{\textbf{BN}}
        & Cont & 89.45 & 62.39  &  89.45 & 62.39 & - & - & 86.48 & 69.08 & 86.19 & 66.26\\
        & Ent & 86.13	& 54.45 & 86.13 & 54.45 & - & - & 79.80 & 62.23 &  71.69 & 56.71 \\ 
        & Neut & 53.50 & 89.29  & 53.50 & 89.29 & - & - & 56.92 & 80.84 &  56.60 & 85.50\\ \cmidrule{2-12}
    
    \multirow{3}{*}{\textbf{EN}} 
        & Cont & 92.45 & 89.40 & - & - & 92.87 & 89.64 & 86.56 & 93.35 & 93.26 & 88.68\\
        & Ent &  88.25 & 86.88 & - & - & 87.84 & 87.84 & 79.26 & 88.08 & 88.20 & 87.25\\ 
        & Neut & 80.02 & 83.90 & - & - & 80.79	& 83.59 & 81.23 & 65.81 & 79.49 & 84.25\\ \cmidrule{2-12} 
        
    \multirow{3}{*}{\textbf{HI}}
        & Cont & 89.95	& 68.62	& 91.08 & 66.65 & 90.78 & 65.45 & 87.98 & 70.54 & - & - \\
        & Ent & 83.00	& 60.24	& 83.12 & 57.49 & 84.60 & 51.32 & 80.02 & 61.62 & - & - \\ 
        & Neut & 56.70	& 85.69	& 54.96 & 86.65 & 53.10 & 88.80 & 57.02 & 81.44 & - & - \\ \cmidrule{2-12}
    
    \multirow{3}{*}{\textbf{UR}}
        & Cont & 67.54	& 79.88	& 82.67 & 64.85 & 84.57 & 62.04 & - & - & 79.22 & 68.50\\
        & Ent & 74.89	& 52.51	& 76.24 & 54.37 & 79.86 & 49.40 & - & - & 79.86 & 49.40\\ 
        & Neut & 56.28	& 62.81	& 53.33 & 80.12 & 51.27 & 84.49 & - & - & 51.27 & 84.49\\ \midrule

\multicolumn{12}{c}{\textbf{Llama 2}} \\ 
\midrule
\multirow{3}{*}{\textbf{BN}}
        & Cont & 06.61 & 06.20  &  14.95 & 06.75 & - & - & 15.28 & 08.47 & 04.82 & 03.01\\
        & Ent & 11.08 & 12.88 & 86.13 & 54.45 & - & - & 26.80 & 53.65 &  22.21 & 41.20 \\ 
        & Neut & 97.01 & 87.27 & 98.57 & 29.56 & - & - & 99.17 & 44.12 &  99.41 & 51.90\\ \cmidrule{2-12}
    
    \multirow{3}{*}{\textbf{EN}} 
        & Cont & 86.78 & 90.00 & - & - & 73.25 & 80.66 & 85.90 & 80.24 & 77.65 & 83.83\\
        & Ent &  52.84 & 95.81 & - & - & 49.05 & 81.98 & 48.87 & 87.01 & 50.69 & 83.77\\ 
        & Neut & 76.00 & 11.38 & - & - & 86.84 & 19.76 & 71.91 & 20.54 & 86.58 & 23.17\\ \cmidrule{2-12} 
        
    \multirow{3}{*}{\textbf{HI}}
        & Cont & 11.88 & 11.80 & 0.60 & 0.60 & 19.10 & 17.54 & 0.89 & 0.89& - & - \\
        & Ent & 13.22 & 15.75 & 0.77 & 0.78 & 20.92 & 29.28 & 0.66 & 0.66 & - & - \\ 
        & Neut & 99.56 & 81.26 & 100.00	& 99.22 & 98.68 & 67.25 & 99.45 & 99.45 & - & - \\ \cmidrule{2-12}
    
    \multirow{3}{*}{\textbf{UR}}
        & Cont & 33.14 & 26.77 & 48.66 & 32.57 & 28.73 & 21.08 & - & - & 10.91 & 11.08\\
        & Ent & 30.50 & 59.88 & 33.17 & 72.93 & 28.56 & 54.79 & - & - & 11.44 & 14.13\\ 
        & Neut & 98.43 & 22.51 & 98.18 & 12.93 & 98.97 & 34.43 &- & - & 99.68 & 74.73\\ \midrule

\multicolumn{12}{c}{\textbf{Gemini}} \\ 
\midrule
\multirow{3}{*}{\textbf{BN}}
        & Cont & 63.32 & 78.47 & 64.60 & 74.11 & - & - & 61.35 & 78.77 & 68.01 & 69.26\\
        & Ent & 66.25 & 54.63 & 67.48 & 46.05 & - & - & 82.42 & 18.40 & 68.56 & 43.59\\ 
        & Neut & 57.12 & 53.49 & 52.59 & 61.54 & - & - & 48.58 & 72.46 & 51.41 & 69.16\\ \cmidrule{2-12}
    
    \multirow{3}{*}{\textbf{EN}} 
        & Cont & 76.97 & 87.84 & - & - & 74.27 & 88.14 & 74.36 & 84.78 & 82.31 & 82.75\\
        & Ent &  73.79 & 71.32 & - & - & 73.53 & 75.51 & 72.78 & 29.94 & 88.20 & 87.25\\ 
        & Neut & 69.44 & 61.95 & - & - & 72.87 & 57.28 & 51.16 & 74.12 & 79.49 & 84.25\\ \cmidrule{2-12} 
        
    \multirow{3}{*}{\textbf{HI}}
        & Cont & 68.03 & 72.87 & 69.40 & 69.94 &  63.55 & 79.34 & 63.83 & 62.99 & - & - \\
        & Ent & 68.92 & 48.74 & 68.24 & 47.60 &  64.16 & 58.86 & 63.57 & 15.57 & - & - \\ 
        & Neut & 52.45 & 64.07 & 52.59 & 68.08 & 57.65 & 48.08 & 43.96 & 77.72 & - & - \\ \cmidrule{2-12}
    
    \multirow{3}{*}{\textbf{UR}}
        & Cont & 42.93 & 71.86 & 59.34 & 77.07 & 64.72 & 73.52 & - & - & 57.93 & 78.37\\
        & Ent & 42.06 & 13.95 & 63.64 & 41.94 & 67.00 & 40.60 & - & - & 66.70 & 41.50\\ 
        & Neut & 63.80 & 63.45 & 50.75 & 52.88 & 50.07 & 63.01 & - & - & 50.61 & 51.89\\ 
\bottomrule
\end{tabular}
}

\caption{Detailed Class-wise result for GPT-4, Llama 2, and Gemini Pro for XNLI dataset. Lang: Language, P: Precision, R: Recall, Cont: Contradiction, Ent: Entailment, Neut: Neutral.} 
\label{tab:detailed-result-gpt-llama-gemini-xnli}
\end{table*}

\begin{table*}[!]
\centering
\resizebox{9.5cm}{!}{
\begin{tabular}{llrrrrrrrrrr}
\toprule
 \multirow{2}{*}{\textbf{Lang.}}& \multirow{2}{*}{\textbf{Class}} & \multicolumn{2}{c}{\textbf{P1}}& \multicolumn{2}{c}{\textbf{P2}}& \multicolumn{2}{c}{\textbf{P3}}& \multicolumn{2}{c}{\textbf{P4}} & \multicolumn{2}{c}{\textbf{P5}}\\ \cmidrule{3-12}
 
 & & \textbf{P.} & \textbf{R.}  & \textbf{P.} & \textbf{R.} & \textbf{P.} & \textbf{R.} & \textbf{P.} & \textbf{R.} & \textbf{P.} & \textbf{R.}  \\
 
\midrule
\multicolumn{12}{c}{\textbf{GPT-4}} \\ 
\midrule

\multirow{7}{*}{\textbf{BN}}
        & Entertainment       & 93.75 & 78.95 & 93.33 & 73.68 & - & - & 92.86 & 68.42 & 87.50 & 73.68\\
        & Geography           & 76.47 & 76.47 & 73.68 & 82.35 & - & - & 77.78 & 82.35 & 73.68 & 82.35 \\ 
        & Health              & 94.44 & 77.27 & 93.33 & 63.64 & - & - & 92.86 & 86.67 & 88.24 & 68.18\\ 
        & Politics            & 93.10 & 90.00 & 93.10 & 90.00 & - & - & 92.86 & 86.67 & 93.10 & 90.00\\
        & Science/Technology  & 90.91 & 98.04 & 85.96 & 96.08 & - & - & 84.75 & 98.04 & 87.50 & 96.08 \\ 
        & Sports              & 91.67 & 88.00 & 88.46 & 92.00 & - & - & 91.30 & 84.00 & 91.30 & 84.00\\ 
        & Travel              & 82.22 & 92.50 & 83.72 & 90.00 & - & - & 80.43 & 92.50 & 84.09 & 92.50\\ 
        
        \cmidrule{2-12}
    \multirow{7}{*}{\textbf{EN}} 
        & Entertainment       & 87.50 & 73.68 & - & - & 93.33 & 73.68 & 93.33 & 73.68 & 93.33 & 73.68\\
        & Geography           & 72.22 & 76.47 & - & - & 72.22 & 76.47 & 72.22 & 76.47 & 72.22 & 76.47 \\ 
        & Health              & 93.33 & 63.64 & - & - & 88.24 & 68.18 & 94.12 & 72.73 & 93.33 & 63.64\\ 
        & Politics            & 93.55 & 96.67 & - & - & 93.55 & 96.67 & 93.55 & 96.67 & 93.55 & 96.67\\
        & Science/Technology  & 84.21 & 94.12 & - & - & 86.21 & 98.04 & 87.72 & 98.04 & 83.33 & 98.04 \\ 
        & Sports              & 91.67 & 88.00 & - & - & 91.30 & 84.00 & 91.67 & 88.00 & 91.67 & 88.00\\ 
        & Travel              & 83.72 & 90.00 & - & - & 88.10 & 92.50 & 88.10 & 88.10 & 90.24 & 92.50\\  
        
        \cmidrule{2-12} 

    \multirow{7}{*}{\textbf{HI}}
        & Entertainment       & 92.31 & 63.16 & 100.00 & 68.42 & 92.31 & 63.16 & 91.67 & 57.89 & - & -\\
        & Geography           & 63.64 & 82.35 & 71.43 & 88.24 & 73.68 & 82.35 & 66.67 & 94.12 & - & - \\ 
        & Health              & 85.00 & 77.27 & 85.71 & 81.82 & 94.12 & 72.73 & 84.21 & 72.73 & - & -\\ 
        & Politics            & 93.33 & 93.33 & 96.55 & 93.33 & 93.33 & 93.33 & 93.10 & 90.00 & - & -\\
        & Science/Technology  & 85.45 & 92.16 & 86.79 & 90.20 & 85.45 & 92.16 & 90.57 & 94.12 & - & - \\ 
        & Sports              & 95.24 & 80.00 & 95.65 & 88.00 & 95.83 & 92.00 & 90.91 & 80.00 & - & -\\ 
        & Travel              & 81.40 & 87.50 & 84.09 & 92.50 & 80.43 & 92.50 & 82.22 & 92.50 & - & -\\  
        \cmidrule{2-12}
   
    \multirow{7}{*}{\textbf{UR}}
        & Entertainment       & 91.67 & 57.89 & 92.31 & 63.16 & 90.91 & 52.63  & - & - & 84.62 & 57.89\\
        & Geography           & 75.00 & 88.24 & 73.33 & 64.71 & 75.00 & 88.24  & - & - & 38.89 & 82.35\\ 
        & Health              & 100.00 & 6364 & 94.44 & 77.27 & 88.24 & 68.18  & - & - & 93.33 & 63.64\\ 
        & Politics            & 93.10 & 90.00 & 96.55 & 93.33 & 93.10 & 90.00 & - & - & 100.00 & 13.33 \\
        & Science/Technology  & 80.65 & 98.04 & 89.29 & 98.04 & 83.33 & 98.04 & - & - & 78.12 & 98.04  \\ 
        & Sports              & 91.67 & 88.00 & 87.50 & 84.00 & 87.50 & 84.00  & - & - & 82.14 & 92.00\\ 
        \midrule

\multicolumn{12}{c}{\textbf{Llama 2}} \\ 
\midrule

\multirow{7}{*}{\textbf{BN}}
        & Entertainment       & 0.00 & 0.00 & 0.00 & 0.00 & - & -  & 0.00 & 0.00 & 0.00 & 0.00\\
        & Geography           & 25.00 & 13.33 & 0.00 & 0.00 & - & -  & 0.00 & 0.00 & 0.00 & 0.00 \\ 
        & Health              & 25.00 & 5.56 & 0.0 & 0.00 & - & -  & 0.00 & 0.00 & 0.00 & 0.00\\ 
        & Politics            & 31.25 & 23.81 & 0.00 & 0.00 & - & -  & 0.00 & 0.00 & 0.00 & 0.00\\
        & Science/Technology  & 27.08 & 74.29 & 04.00 & 10.00 & - & -  & 26.71 & 91.49 & 25.29 & 52.38 \\ 
        & Sports              & 33.33 & 16.67 & 0.00 & 0.00 & - & - & 0.00 & 0.00\\ 
        & Travel              & 09.09 & 03.57 & 0.00 & 0.00 & - & -  & 0.00 & 0.00\\  
        \cmidrule{2-12}
    \multirow{7}{*}{\textbf{EN}} 
        & Entertainment       & 76.92 & 55.56 & - & - & 77.78 & 77.78 & 83.33 & 26.32 & 69.23 & 75.00 \\
        & Geography           & 33.33 & 58.82 &  - & - & 36.67 & 64.71 & 33.33 & 35.29 & 35.71 & 66.67 \\ 
        & Health              & 81.82 & 42.86 & - & - & 85.71 & 54.55 & 80.00 & 18.18 & 92.31 & 63.16 \\ 
        & Politics            & 68.97 & 68.97 & - & - & 71.05 & 93.10 & 100.00 & 40.00 & 71.88 & 92.00 \\
        & Science/Technology  & 51.76 & 86.27 &  - & - & 73.13 & 96.08 & 36.69 & 100.00 & 74.55 & 82.00\\ 
        & Sports              & 92.86 & 52.00 & - & - & 91.30 & 84.00 & 100.00 & 48.00 & 94.44 & 85.00\\ 
        & Travel              & 69.23 & 26.47 & - & - & 100.00 & 17.65 & 100.00 & 30.00 & 57.14 & 16.00\\        \cmidrule{2-12} 

    \multirow{7}{*}{\textbf{HI}}
       & Entertainment       & 25.00 & 05.26 & 33.33 & 05.56 & 100.0 & 05.56 & 0.00 & 0.0 & - & -\\
        & Geography           & 0 & 0 & 25.00 & 05.88 & 0.00 & 0.00 & 00.0 & 0.00 & - & - \\ 
        & Health              & 0 & 0 & 40.0 & 18.18 & 0.00 & 0.00 & 100.00 & 04.55 & - & -\\ 
        & Politics            & 0 & 0 & 50.00 & 100.00 & 50.00 & 03.33 & 24.32 & 88.24 & - & -\\
        & Science/Technology  & 25.26 & 94.12 & 29.81 & 94.12 & 26.18 & 98.04 & 24.32 & 88.24 & - & - \\ 
        & Sports              & 0 & 0 & 33.33 & 04.00 & 0 & 0 & 0.00 & 0.00 & - & -\\ 
        & Travel              & 40.00 & 05.00 & 081.25 & 32.50 & 75.00 & 07.5 & 22.22 & 05.00 & - & -\\ 
        \cmidrule{2-12}
    
    \multirow{7}{*}{\textbf{UR}}
        & Entertainment       & 0.00 & 0.00 & 0.00 & 0.00 & 0.00 & 0.00 & - & - & 0.00 & 0.00\\
        & Geography           & 0.00 & 0.00 & 0.00 & 0.00 & 0.00 & 0.00 & - & - & 0.00 & 0.00 \\ 
        & Health              & 0.00 & 0.00 & 0.00 & 0.00 & 0.00 & 0.00 & - & - & 0.00 & 0.00\\ 
        & Politics            & 0.00 & 0.00 & 0.00 & 0.00 & 0.00 & 0.00 & - & - & 0.00 & 0.00\\
        & Science/Technology  & 23.08 & 81.82 & 25.38 & 98.04 & 24.62 & 96.00 & - & - & 25.00 & 96.00 \\ 
        & Sports              & 0.00 & 0.00 & 0.00 & 0.00 & 0.00 & 0.00 & - & - & 0.00 & 0.00\\ 
        & Travel              & 0.00 & 0.00 & 0.00 & 0.00 & 0.00 & 0.00 & - & - & 0.00 & 0.00\\ 

        \midrule

\multicolumn{12}{c}{\textbf{Gemini}} \\ 
\midrule

\multirow{7}{*}{\textbf{BN}}
        & Entertainment       & 85.71 & 63.16 & 87.50 & 73.68 & - & - & 73.68 & 73.68 & 83.33 & 78.95\\
        & Geography           & 76.47 & 76.47 & 73.68 & 82.35 & - & - & 70.59 & 70.59 & 68.75 & 64.71 \\ 
        & Health              & 100.00 & 72.73 & 88.24 & 68.18 & -& - & 83.33 & 68.18 & 88.24 & 68.18\\ 
        & Politics            & 96.67 & 96.67 & 93.10 & 90.00 & - & - & 93.33 & 93.33 & 87.88 & 96.67\\
        & Science/Technology  & 77.97 & 90.20 & 81.67 & 96.08 & - & - & 91.49 & 84.31 & 85.71 & 94.12 \\ 
        & Sports              & 88.89 & 96.00 & 84.62 & 88.00 & - & - & 91.30 & 84.00 & 88.00 & 88.00\\ 
        & Travel              & 80.49 & 82.50 & 89.19 & 82.50 & - & - & 70.00 & 87.50 & 84.62 & 82.50\\ 
        \cmidrule{2-12}

    \multirow{7}{*}{\textbf{EN}} 
        & Entertainment       & 100.00 & 68.42 & - & - & 93.33 & 73.68 & 100.00 & 63.16 & 100.00 & 68.42\\
        & Geography           & 83.33 & 58.82 & - & - & 84.62 & 64.71 & 83.33 & 58.82 & 71.43 & 58.82 \\ 
        & Health              & 93.33 & 63.64 & - & - & 86.67 & 59.09 & 85.71 & 54.55 & 94.12 & 7273\\ 
        & Politics            & 84.85 & 93.33 & - & - & 82.35 & 93.33 & 87.50 & 93.33 & 78.38 & 96.67\\
        & Science/Technology  & 76.12 & 100.00 & - & - & 77.27 & 100.00 & 72.86 & 100.00 & 80.65 & 98.04 \\ 
        & Sports              & 85.71 & 96.00 & - & - & 84.62 & 88.00 & 82.14 & 92.00 & 88.89 & 96.00\\ 
        & Travel              & 91.67 & 82.50 & - & - & 91.43 & 80.00 & 91.67 & 82.50 & 94.12 & 80.00\\  \cmidrule{2-12}

    \multirow{7}{*}{\textbf{HI}}
        & Entertainment       & 86.67 & 68.42 & 92.86 & 68.42 & 69.23 & 47.37 & 85.71 & 63.16 & - & -\\
        & Geography           & 61.54 & 47.06 & 81.25 & 76.47 & 71.43 & 58.82 & 83.33 & 58.82 & - & - \\
        & Health              & 93.75 & 68.18 & 100.00 & 77.27 & 100.00 & 72.73 & 100.00 & 68.180 & - & -\\ 
        & Politics            & 87.50 & 93.33 & 87.10 & 90.0 & 84.85 & 93.33 & 84.85 & 93.33 & - & -\\
        & Science/Technology  & 84.75 & 98.04 & 82.26 & 100.00 & 81.03 & 92.16 & 78.69 & 94.12 & - & - \\ 
        & Sports              & 82.14 & 92.00 & 85.71 & 96.00 & 88.46 & 92.00 & 88.89 & 96.00 & - & -\\ 
        & Travel              & 80.49 & 82.50 & 91.67 & 82.50 & 79.55 & 87.50 & 80.95 & 85.00 & - & -\\  \cmidrule{2-12}
    \multirow{7}{*}{\textbf{UR}}
        & Entertainment       & 100.00 & 42.11 & 85.71 & 63.16 & 92.31 & 63.16 & - & - & 71.43 & 52.63\\
        & Geography           & 58.82 & 85.82 & 64.71 & 64.71 & 64.71 & 64.71 & - & - & 47.83 & 64.71\\ 
        & Health              & 100.00 & 68.18 & 93.33 & 63.64 & 100.00 & 77.71 & - & - & 80.95 & 77.27\\ 
        & Politics            & 87.10 & 90.00 & 84.38 & 63.64 & 92.32 & 93.33 & - & - & 83.33 & 16.67\\
        & Science/Technology  & 76.56 & 96.08 & 75.38 & 96.08 & 84.75 & 98.04 & - & - & 76.67 & 90.20 \\ 
        & Sports              & 88.46 & 92.00 & 85.19 & 92.00 & 88.46 & 92.00 & - & - & 67.65 & 92.00\\ 
        & Travel              & 79.07 & 85.00 & 88.24 & 75.00 & 87.80 & 0.9 & - & - & 80.43 & 92.50\\  
\bottomrule
\end{tabular}
}

\caption{Detailed Class-wise result for GPT-4, Llama 2, and Gemini Pro for SIB-200 dataset. Lang: Language, P: Precision, R: Recall, Cont: Contradiction, Ent: Entailment, Neut: Neutral.} 
\label{tab:detailed-result-gpt-llama-gemini-sib}
\end{table*}

% \begin{table}[!ht]
% \centering
% \resizebox{0.9\linewidth}{!}{
% \begin{tabular}{llc|c|c|c|c}
% \toprule
% \multirow{2}{*}{\textbf{Model}} & \multicolumn{2}{c}{\textbf{Setting P1}}& \textbf{Setting P2}& \textbf{Setting P3}& \textbf{Setting P4} & \textbf{Setting P5}\\ 
% \cmidrule{2-7} 
% &\textbf{Lang.} & \textbf{Miss} & \textbf{Miss.} & \textbf{Miss.} & \textbf{Miss.} & \textbf{Miss.} \\ \midrule

% \multirow{4}{*}{\textbf{GPT-4}} 
%         & \textbf{BN} & 1 &  13 & - & 2 & 48\\
%         &\textbf{EN} & 0 &  - &  0 & 0 & 0\\
%         &\textbf{HI} & 9 &  13 &  17  & 7 & -\\ 
%         &\textbf{UR} & 43 &  53 &  18  & - & 130\\ \midrule

% \multirow{4}{*}{\textbf{Lllama 2}} 
%         & \textbf{BN} & 4147 & 1678 & - & 2195 & 2827\\
%         &\textbf{EN} & 19 & - & 619 & 322 & 588\\
%         &\textbf{HI} & 4010 & 4966 & 3306 & 4923  & -\\ 
%         &\textbf{UR} & 1308 & 790 & 1917 & - & 3872 \\ \midrule

% \multirow{4}{*}{\textbf{Gemini}} 
%         & \textbf{BN} & 142 & 149 - & 68 & 127\\
%         &\textbf{EN} & 138 & - & 143 & 80 & 129\\
%         &\textbf{HI} & 111 & 131  & 138 & 133 & -\\ 
%         &\textbf{UR} & 1330 & 105 & 85 & - & 77\\ 

% \bottomrule
% \end{tabular}
% }
% \caption{Invalid label (i.e. Miss-classification) returned by LLMs across the settings, models, and languages. Lang.: language, Miss.: Miss-classification.}
% \label{tab: miss-classification}
% \end{table}

\begin{table}[!t]
\centering
% \resizebox{\linewidth}{!}{
\scalebox{0.60}{
\begin{tabular}{llrrrrr}
\toprule
\multirow{1}{*}{\textbf{Model}} & \multirow{1}{*}{\textbf{Lang.}} &\multicolumn{1}{c}{\textbf{ P1}} & \textbf{ P2}& \textbf{ P3}& \textbf{ P4} & \textbf{ P5}\\ 
%\cmidrule{3-7} 
%&& \multicolumn{1}{c}{\textbf{Miss}} & \multicolumn{1}{c}{\textbf{Miss.}} & \multicolumn{1}{c}{\textbf{Miss.}} & \multicolumn{1}{c}{\textbf{Miss.}} & \multicolumn{1}{c}{\textbf{Miss.}} \\ 
\midrule

\multirow{4}{*}{\textbf{GPT-4}} 
        & \textbf{BN} & 1 &  13 & - & 2 & 48 \\
        &\textbf{EN} & 0 &  - &  0 & 0 & 0 \\
        &\textbf{HI} & 9 &  13 &  17  & 7 & -\\ 
        &\textbf{UR} & 43 &  53 &  18  & - & 130\\ \midrule

\multirow{4}{*}{\textbf{Llama 2}} 
        &\textbf{BN} & 4147 & 1678 & - & 2195 & 2827\\
        &\textbf{EN} & 19 & - & 619 & 322 & 588\\
        &\textbf{HI} & 4010 & 4966 & 3306 & 4923  & -\\ 
        &\textbf{UR} & 1308 & 790 & 1917 & - & 3872 \\ \midrule

\multirow{4}{*}{\textbf{Gemini}} 
        &\textbf{BN} & 142 & 149 & - & 68 & 127\\
        &\textbf{EN} & 138 & - & 143 & 80 & 129\\
        &\textbf{HI} & 111 & 131  & 138 & 133 & -\\ 
        &\textbf{UR} & 1330 & 105 & 85 & - & 77\\ 

\bottomrule
\end{tabular}
}
\caption{Number of invalid labels returned by LLMs across the settings, models, and languages for the XNLI dataset. Lang.: language.}
\label{tab: miss-classification-xnli}
\end{table}

\begin{table}[!t]
\centering
% \resizebox{\linewidth}{!}{
\scalebox{0.70}{
\begin{tabular}{llrrrrr}
\toprule
\multirow{1}{*}{\textbf{Model}} & \multirow{1}{*}{\textbf{Lang.}} &\multicolumn{1}{c}{\textbf{ P1}} & \textbf{ P2}& \textbf{ P3}& \textbf{ P4} & \textbf{ P5}\\ 
%\cmidrule{3-7} 
%&& \multicolumn{1}{c}{\textbf{Miss}} & \multicolumn{1}{c}{\textbf{Miss.}} & \multicolumn{1}{c}{\textbf{Miss.}} & \multicolumn{1}{c}{\textbf{Miss.}} & \multicolumn{1}{c}{\textbf{Miss.}} \\ 
\midrule

\multirow{4}{*}{\textbf{GPT-4}} 
        & \textbf{BN} & 0 &  1 & - & 0 & 0 \\
        &\textbf{EN} & 0 &  - &  0 & 2 & 1 \\
        &\textbf{HI} & 1 &  0 &  1  & 2 & -\\ 
        &\textbf{UR} & 0 &  1 &  1  & - & 3\\ \midrule

\multirow{4}{*}{\textbf{Llama 2}} 
        &\textbf{BN} & 12 & 0 & - & 0 & 15\\
        &\textbf{EN} & 46 & - &6 & 17 & 95\\
        &\textbf{HI} & 12 & 13 & 7 & 14 & -\\ 
        &\textbf{UR} & 27 & 3 & 8 & - & 11 \\ \midrule

\multirow{4}{*}{\textbf{Gemini}} 
        &\textbf{BN} & 1 & 2 & - & 2 & 3\\
        &\textbf{EN} & 4 & - & 2 & 8 & 2\\
        &\textbf{HI} & 0 & 1 & 3 & 3 & -\\ 
        &\textbf{UR} & 3 & 2 & 3 & - & 12\\ 

\bottomrule
\end{tabular}
}
\caption{Number of invalid labels returned by LLMs across the settings, models, and languages for the SIB-200 dataset. Lang.: language.}
\label{tab: miss-classification-sib}
\end{table}

\end{document}